\documentclass[10pt,twocolumn,letterpaper]{article}

\usepackage{cvpr}              %
\usepackage{xspace}
\usepackage{comment}
\usepackage{graphicx}
\usepackage{pgf}
\usepackage{stfloats}
\usepackage{xcolor}
\usepackage{colortbl} %
\usepackage{multirow}
\graphicspath{{figures/}}
\usepackage{minted}
\usepackage{listings}
\usepackage{xcolor}

\newcommand{\modelname}{MATCH\xspace}
\newcommand{\modelnamelong}{\modelname (Multi-view Avatars from Topologically Corresponding Heads)\xspace}

\renewcommand{\paragraph}[1]{\noindent\textbf{#1}}

\definecolor{cvprblue}{rgb}{0.21,0.49,0.74}
\definecolor{tabfirst}{rgb}{1, 0.7, 0.7} %
\definecolor{tabsecond}{rgb}{1, 0.85, 0.7} %
\definecolor{tabthird}{rgb}{1, 1, 1} %

\usepackage[pagebackref,breaklinks,colorlinks,allcolors=cvprblue]{hyperref}

\def\paperID{****} %
\def\confName{CVPR}
\def\confYear{2026}

\title{Feed-forward Gaussian Registration for Head Avatar Creation and Editing}

\author{
Malte Prinzler$^{1,2\footnotemark}$
~~~~
Paulo Gotardo$^{2}$
~~~~
Siyu Tang$^{1}$
~~~~
Timo Bolkart$^{2}$
\vspace{0.2cm}
\\
$^1$ETH Zürich 
~~~~
$^2$Google
}

\begin{document}
\twocolumn[{%
\renewcommand\twocolumn[1][]{#1}%
\maketitle
\begin{center}
    \vspace{-0.5cm}
    \centering    
    \captionsetup{type=figure}
    \includegraphics[width=0.99\textwidth]{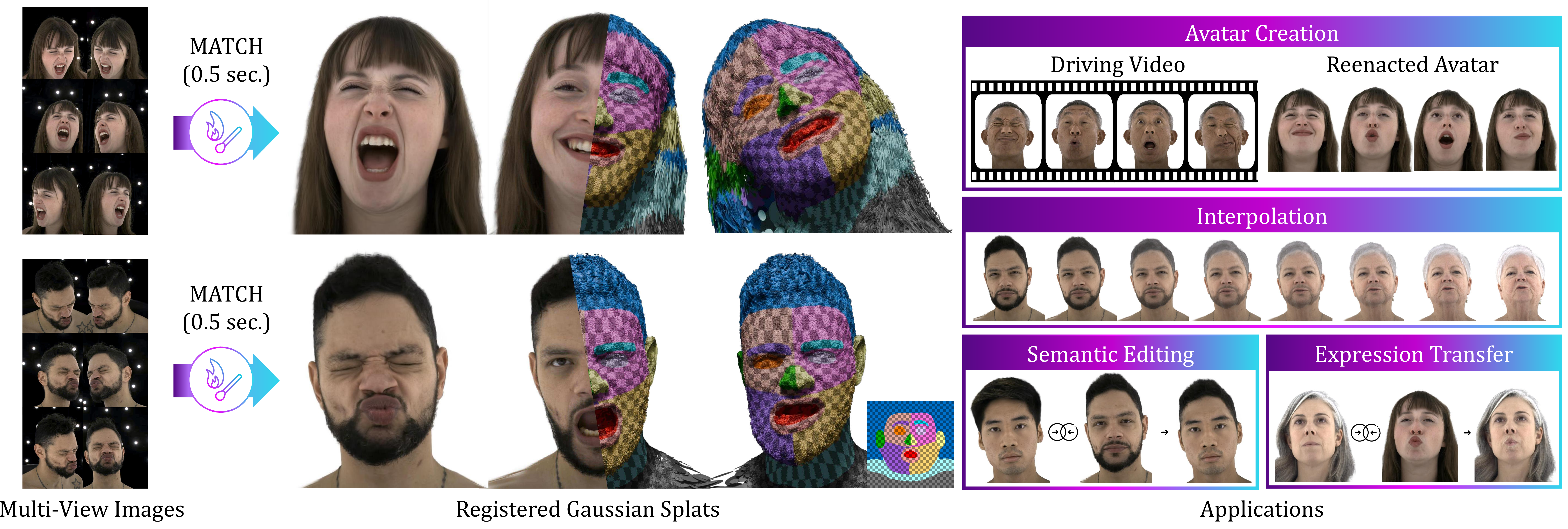}
    \caption{
    Given calibrated multi-view images as input, \modelname infers static Gaussian splat textures in 0.5 seconds. 
    The resulting Gaussians are in dense semantic correspondence across subjects and expressions.
    This enables diverse downstream applications such as fast head avatar creation, interpolation, semantic editing, and expression transfer.
    For visualization, we show 6 of the 12 input images, display predicted Gaussians for three separate frames, and apply a checkerboard semantic texture to highlight the dense correspondence.
    }
    \label{fig:teaser}
    \vspace{-0.01cm}
\end{center}%
}]
\begingroup
\renewcommand{\thefootnote}{\fnsymbol{footnote}}
\footnotetext[1]{Work done while Malte Prinzler was an intern at Google.}
\endgroup

\begin{abstract}
We present \modelnamelong, a multi-view Gaussian registration method for high-quality head avatar creation and editing.
State-of-the-art multi-view head avatars require time-consuming head tracking followed by expensive avatar optimization, resulting in creation times exceeding one day.
In contrast, \modelname directly predicts Gaussian splats in correspondence from calibrated multi-view images in 0.5 seconds per frame. 
While the learned intra-subject correspondence across frames allows us to quickly build personalized head avatars, correspondence across subjects enables expression transfer, optimization-free tracking, semantic editing, and interpolation.
We establish these correspondences with a transformer that predicts textures of Gaussian splats.
To this end, we introduce a novel attention block, in which each UV map token attends exclusively to image tokens depicting its corresponding mesh region. 
\modelname outperforms existing methods for novel-view synthesis, geometry registration, and head avatar generation, the latter being $\mathit{10\times}$ faster than the qualitatively closest baseline.
The code and model weights are available on the \href{https://malteprinzler.github.io/projects/match/}{project website}.

\end{abstract}
    
\vspace{-0.5cm}
\section{Introduction}
\label{sec:intro}

The growing demand for realistic digital humans in telepresence, film, and gaming has intensified the need for scalable methods that can rapidly create controllable, photo-realistic head avatars.
Obtaining state-of-the-art head avatars from multi-view in-studio captures typically relies on a two-stage pipeline.
This involves first establishing cross-view and temporal correspondence via mesh-based tracking before optimizing surface-attached primitives, such as Gaussian splats~\cite{kerbl20233d, huang20242d}, to model appearance~\cite{giebenhain2024npga, qian2024gaussianavatars, zielonka2025gem}.
Despite its high-fidelity results, this two-stage process is a major computational bottleneck, as creating a single personalized avatar typically requires hours, or even days, of optimization.
Consequently, scaling to a large number of subjects becomes prohibitively expensive.

Our method \modelname overcomes these limitations by directly predicting Gaussian splats in dense semantic correspondence from multi-view images in $0.5$ seconds per frame. 
In this context, dense semantic correspondence means that all output Gaussian textures share a fixed topology in which a specific Gaussian consistently represents the same semantic region (e.g., the nose tip), regardless of identity or expression.
Crucially, this correspondence enables downstream applications such as building lightweight head avatars in the form of linear Gaussian Eigen Models (GEM)~\cite{zielonka2025gem}, or performing direct expression transfer and semantic editing (see Figure~\ref{fig:teaser}).
By bypassing both time-consuming mesh tracking (10.7 hours) and animatable Gaussians optimization (27.7 hours), our approach reduces the total GEM avatar creation time by a factor of 10.

\modelname employs a transformer architecture inspired by Large Reconstruction Models (LRM) \cite{tang2024lgm, xu2024grm, zhang2024gs}. 
The UV texture map and input images are tokenized and processed by a sequence of transformer blocks. 
However, na{\" i}vely attending to all image and UV tokens \cite{wang2024dust3r,wang2025vggt,he2025lam,wu2025fastavatar} is computationally expensive and leads to poor generalization on unseen subjects \cite{kirschstein2025avat3r}.
Instead, we introduce a novel registration-guided attention block.
It restricts each UV token to attend only to image tokens displaying the relevant head region, which reduces compute complexity while simultaneously improving synthesis quality. %

In summary, \modelname represents heads with Gaussian textures in dense semantic correspondence that are regressed directly from multi-view input images using a novel registration-guided attention mechanism.
These textures can subsequently be processed into lightweight GEM~\cite{zielonka2025gem} avatars, drivable by monocular videos.
Finally, we show that dense Gaussian correspondence enables further applications, such as semantic editing and expression transfer.

\section{Related Work}
\label{sec:related_work}
 
\paragraph{Head correspondences.}
The concept of correspondence across head captures was introduced by early methods such as Blanz and Vetter~\cite{blanz19993dmm} two decades ago, followed by extensive research on the optimization-based alignment of a common mesh topology to unstructured 3D head scans \cite{blanz19993dmm, amberg2008expression, hutton2001dense, passalis2011using, mpiperis2008bilinear, salazar2014fully, li2017flame}.
Recent learning-based approaches have moved towards predicting registered meshes directly from the raw scans \cite{bahri2021shape, liu20193d, zheng2022imface}, calibrated multi-view input images \cite{bolkart2023tempeh, li2021tofu, li2024grape, liu2022refa, filntisis2026mochi}, or even monocular in-the-wild images \cite{giebenhain2025pixel3dmm}.  
While these works are restricted to geometry reconstructions, our method infers registered Gaussian splats representing both geometry and appearance. 

Registered head scans enable the construction of 3D morphable models (3DMM)~\cite{blanz19993dmm}, which are used to establish temporal and spatial correspondences across single- or multi-view video frames through subject-specific head tracking \cite{qian2024vhap, taubner2024flowface, li2024topo4d}.
Our method allows us to avoid this slow optimization process (12 s/frame for VHAP~\cite{qian2024vhap}) by directly predicting registered Gaussians in 0.5~s per frame. 

\paragraph{Optimization-based head avatars.}
Current personalized head avatars require 3DMM tracking to establish correspondences before avatar optimization.
These methods attach appearance representations, such as RGB textures \cite{grassal2022neural}, localized radiance field primitives \cite{lombardi2021mixture}, or Gaussian splats \cite{giebenhain2024npga, xu2023gaussianheadavatar, zielonka2025gem,li2024uravatar}, to the tracked geometry and optimize them against training images.
GEM~\cite{zielonka2025gem} shows that such high-quality avatars can be distilled into lightweight PCA-based representations.
The high quality of the reconstructed avatars comes at the cost of long optimization times (45~h per avatar for GEM).
We show that \modelname's predictions can be used to significantly accelerate the reconstruction of lightweight head avatars to 4.6~h per avatar.

\paragraph{Inference-based head avatars.}
One way to avoid the compute-heavy registration process is to infer animatable head avatars directly from one or a few images, typically at the cost of inferior synthesis quality. 
While early methods in this field focused on direct 2D image generation \cite{siarohin2019first, doukas2021headgan, burkov2020neural}, recent works have moved towards 3D representations such as texturized meshes \cite{khakhulin2022realistic}, Neural Radiance Fields~\cite{nerface_Gafni_2021_CVPR, prinzler2022diner, prinzler2025joker}, animatable triplanes \cite{tran2024voodoo, drobyshev2022megaportraits, drobyshev2024emoportraits, deng2024portrait4d, chu2024gpavatar}, or 3DMM-attached Gaussian splats \cite{he2025lam, li2025panolam, wu2025fastavatar}. 
Similar to our approach, LAM~\cite{he2025lam} and FastAvatar~\cite{wu2025fastavatar} estimate Gaussians with a fixed UV location on a template mesh. However, their Gaussians are predicted in an unposed canonical space, while our method reconstructs the observed expression as-is in posed space with improved fidelity.
Avat3r~\cite{kirschstein2025avat3r} and Facelift~\cite{lyu2025facelift} predict pixel-aligned Gaussians~\cite{xu2024grm, zhang2024gs, tang2024lgm, lin2025diffsplat}. 
Unlike our method, these do not exhibit any cross-frame correspondences.

\section{Method}
\begin{figure*}[t]
    \centering
    \includegraphics[width=\linewidth]{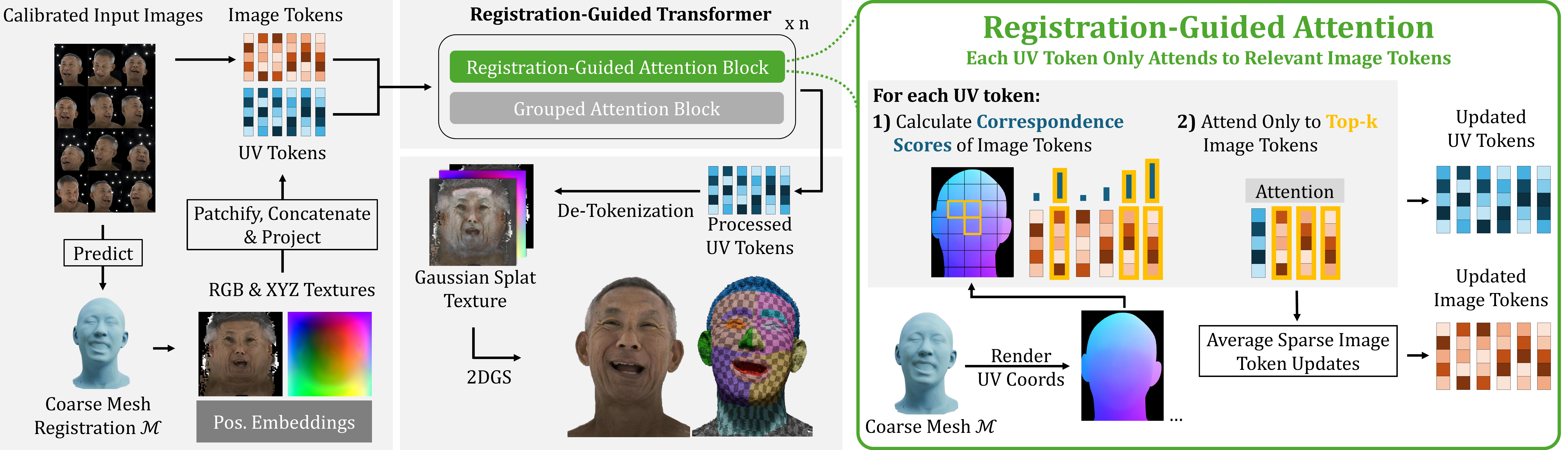}
    \caption{\textbf{Overview}. Given calibrated multi-view input images, \modelname first predicts a coarse mesh registration using a pretrained network. We obtain RGB and XYZ textures combined with learnable positional embeddings to encode UV tokens and follow GS-LRM~\cite{zhang2024gs} to tokenize the input images.
    The image and UV tokens serve as input to a transformer with two alternating attention blocks.
    In the novel registration-guided attention block, we render UV coordinate images from the input views, and for each UV token restrict the attention to image tokens displaying the relevant mesh region.
    The subsequent grouped attention block performs attention across the UV tokens and the tokens of each input image separately. 
    The transformer outputs processed UV tokens that are projected into a texture of Gaussians.}
    \label{fig:method}
\end{figure*}

This section presents \modelname, a feed-forward method for reconstructing photorealistic 3D heads as textures of registered Gaussian splats from calibrated multi-view images.
We first describe the components of \modelname and then show how it can be applied to speed up the reconstruction process of lightweight, subject-specific head avatars (\Cref{sec:avatar_creation}).

\subsection{\modelname}
Given $V$ multi-view input images $\mathcal{I}\in\mathbb{R}^{H_\text{img}\times W_\text{img}\times3}$ with known camera parameters, \modelname predicts a UV texture of Gaussian splat attributes $\mathcal{G}_{\{ \mathbf{c}, \boldsymbol{\alpha}, \boldsymbol{\phi}, \boldsymbol{\sigma}, \boldsymbol{\theta}\}} \
\in\mathbb{R}^{H_\text{uv}\times W_\text{uv}\times C}$.
Every texel is a $C$-vector that encodes RGB color $\mathbf{c}$, opacity $\boldsymbol{\alpha}$, location $\boldsymbol{\phi}$, anisotropic scale $\boldsymbol{\sigma}$, and rotation quaternion $\boldsymbol{\theta}$ of a Gaussian splat.
An overview of \modelname's pipeline is presented in \Cref{fig:method}.

\paragraph{Image tokenization.}
Following recent works \cite{zhang2024gs, kirschstein2025avat3r}, we convert the high-resolution images $\mathcal{I}$ into image tokens $\mathcal{T}_\text{img}$ within low-resolution grids $\in\mathbb{R}^{H'_\text{img} \times W'_\text{img}\times d}$ that can be processed by the transformer.
First, the images are concatenated with Pl{\" u}cker ray coordinates~\cite{plucker1865xvii}.
The result is split into patches and converted into tokens using a 2D convolutional layer with stride and kernel size equal to the patch size $p_\text{img}$.
Following Avat3r~\cite{kirschstein2025avat3r}, we fuse the image tokens with Sapiens~\cite{khirodkar2024sapiens} features through concatenation and linear projection, which yields the final tokens of dimension $d$.

\paragraph{Coarse mesh registration.}
Before computing the UV tokens, %
TEMPEH's~\cite{bolkart2023tempeh} global stage without head localization is used to estimate a coarse mesh $\mathcal{M}\in\mathbb{R}^{N_\text{vert}\times3}$ from the input images.
TEMPEH is trained against the ground-truth vertices of the Ava-256 dataset~\cite{martinez2024codec}, including hair proxy geometry, and adopts the provided mesh topology.

\paragraph{UV tokenization.}
During UV tokenization, a high-resolution UV texture is encoded as a set of UV tokens $\mathcal{T}_\text{uv}$ in a low-resolution grid $\in\mathbb{R}^{H'_\text{uv} \times W'_\text{uv}\times d}$ that can be processed by the transformer. 
We first calculate a dense 3D location texture through per-texel barycentric interpolation of $\mathcal{M}$'s vertex locations.
Second, we obtain an RGB texture by reprojecting the input images onto $\mathcal{M}$. 
These RGB and 3D location textures are concatenated and divided into non-overlapping patches of size $p_\text{uv}$. 
Finally, they are flattened, concatenated with learnable positional embeddings, and linearly projected to yield $d$-dimensional UV tokens.

\paragraph{Registration-guided attention.}
\begin{figure}[t]
    \centering
    \includegraphics[width=\linewidth]{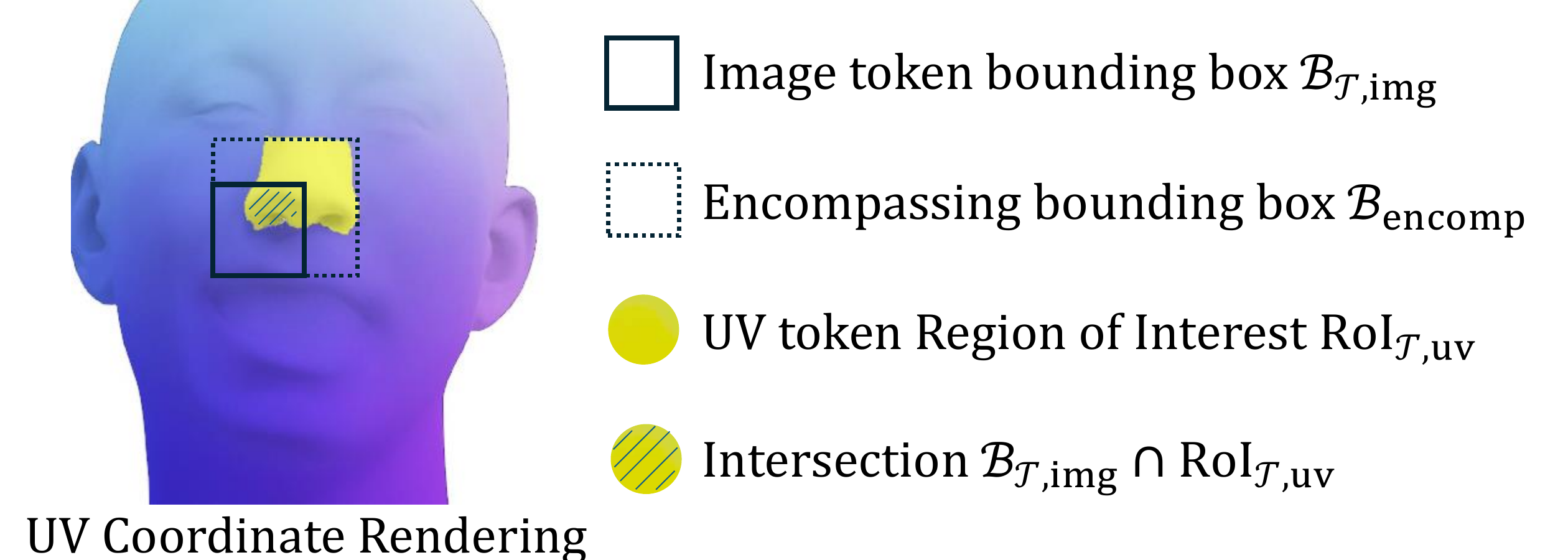}
    \caption{Correspondence score estimation between image tokens and UV tokens. To ease visualization, the full mesh is rasterized in overlay with the UV renders and patch sizes are increased.}
    \label{fig:correspondence}
\end{figure}
The image and UV tokens are processed by a transformer that contains alternating blocks of registration-guided attention and grouped attention. 
The registration-guided attention blocks constrain each UV token to only attend to image tokens of the corresponding head region (\Cref{fig:method} right).
To determine which tokens should attend to each other, we estimate a correspondence score for each pair of UV and image tokens:
We use the coarse mesh $\mathcal{M}$ to rasterize UV coordinates onto the input image planes. Since each UV token represents a texture patch, we can filter the rasterized UV coordinates by the respective coordinate range to obtain a region of interest $\text{RoI}_{\mathcal{T}, \text{uv}}$ for that UV token, shown in yellow in \Cref{fig:correspondence}.
Now let $\mathcal{B}_{\mathcal{T},\text{img}}$ denote the bounding box of a particular image token, and $\mathcal{B}_\text{encomp}$ the bounding box that includes both $\mathcal{B}_{\mathcal{T},\text{img}}$ and $\text{RoI}_{\mathcal{T},\text{uv}}$. 
The correspondence score $S$ between the UV- and image token is defined as:
\begin{equation}
    S\left(\mathcal{T}_\text{uv}, \mathcal{T}_\text{img}\right) = \frac{\text{RoI}_{\mathcal{T},\text{uv}} \cap \mathcal{B}_{\mathcal{T},\text{img}}}{\mathcal{B}_{\mathcal{T},\text{img}}} + \lambda \cdot \frac{\text{RoI}_{\mathcal{T},\text{uv}}}{\mathcal{B}_\text{encomp}},
\end{equation}
where the arithmetic is performed on the respective areas in pixels. 
The first term measures the ratio of pixels within $\mathcal{B}_{\mathcal{T},\text{img}}$ coinciding with $\text{RoI}_{\mathcal{T},\text{uv}}$, while the second term accounts for image tokens in $\text{RoI}_{\mathcal{T},\text{uv}}$'s vicinity (weighted by $\lambda = 0.1$).
After evaluating $S(\cdot)$ across all image tokens, for each UV token, we only attend to the $k_{\mathcal{T},\text{img}}$ highest-scoring image tokens. 
This localizes the reconstruction task, which improves generalization, and keeps the attention context length constant with an increasing number of input images, reducing computation costs for high image counts.
After the attention operation, we obtain updated UV tokens, but only sparsely and potentially redundantly updated image tokens, which we address by averaging over all their occurrences.

\paragraph{Grouped attention.}
This block performs attention separately on the UV tokens and the image tokens of each input image, propagating information to unobserved head regions and enabling image-space feature processing with compute complexity scaling linearly in the number of input images.

\paragraph{UV de-tokenization.}
The transformer outputs processed UV tokens. 
We linearly project each token to a texture patch of Gaussian parameters with shape $p_\text{uv} \times p_\text{uv} \times C$. 
These are assembled to the output Gaussian splat texture of shape $H_\text{uv}\times W_\text{uv} \times C$. 
For color and location, we follow Avat3r's~\cite{kirschstein2025avat3r} skip connections and apply the predictions as offsets to the initial values obtained during UV tokenization.

\paragraph{Training.}
Supervised by multi-view images and ground-truth meshes, \modelname is trained by minimizing:
\begin{equation}
    \mathcal{L}_\text{total} = \mathcal{L}_\text{photometric}  + w_\text{geometry}\cdot\mathcal{L}_\text{geometry} + w_\text{reg}\cdot\mathcal{L}_\text{reg}.
\end{equation}
The photometric loss $\mathcal{L}_\text{photometric}$ matches the predicted appearance against the ground-truth images,
\begin{equation}
    \mathcal{L}_\text{photometric} = w_\text{LPIPS} \cdot \mathcal{L}_\text{LPIPS} +  w_\text{L1}  \cdot \mathcal{L}_\text{L1} +  w_\text{SSIM} \cdot \mathcal{L}_\text{SSIM}
\end{equation}
with perceptual loss \cite{zhang2018unreasonable} $\mathcal{L}_\text{LPIPS}$ and Structural Similarity Index Measure (SSIM)~\cite{wang2004image} $\mathcal{L}_\text{SSIM}$.
We mask out the torso and upper shoulders using a pretrained semantic segmentation model \cite{khirodkar2024sapiens}.
$\mathcal{L}_\text{geometry}$ minimizes the L2-norm between the predicted Gaussian 3D locations and dense target locations from the ground-truth mesh registration. 
The regularization loss $\mathcal{L}_\text{reg}$ applies an L2 loss between the predictions and predefined target values for scale and opacity. 

\paragraph{Implementation details:}
Our training consists of three stages. 
First, we train solely on the Ava-256 dataset~\cite{martinez2024codec}, using $\mathcal{L}_\text{geometry}$ and $\mathcal{L}_\text{reg}$ only. 
In the second stage, we add $\mathcal{L}_\text{photometric}$. 
In the last stage, we train on a combination of the Ava-256 and the NeRSemble~v2 dataset~\cite{kirschstein2023nersemble}. 
Since NeRSemble does not provide ground-truth geometry annotations, we deactivate $\mathcal{L}_\text{geometry}$ on samples drawn from it. 

We adopt Ava-256's mesh topology and UV texture layout. 
\modelname predicts 64$\times$64 UV tokens, each corresponding to 16$\times$16 texture patches, totalling a 1024$\times$1024 texture with 1M Gaussians that can be rendered at 570 fps~\cite{huang20242d}. 
The input images are divided into $8\times8$ patches.

Each training sample contains 12 head-centered images with a resolution of $640 \times 512$, which are used both as model input and rendering target. 
We train \modelname for 860k iterations on 4 NVIDIA H100 80GB GPUs, with per-GPU batch size 1, taking 11.8 days. 
Please see \Cref{sec:supp_implementation} for details.

\subsection{Creation of Subject-Specific Head Avatars}
\label{sec:avatar_creation}
Practical applications require animatable avatars.
GEM~\cite{zielonka2025gem} demonstrates the distillation of high-quality 3D head avatars (computationally and memory expensive) into a lightweight representation.
We adapt this procedure to obtain controllable head avatars from sequences of Gaussian textures predicted by \modelname.

\paragraph{GEM.}
Given a multi-view video sequence, GEM performs mesh-based head tracking and optimizes a CNN-based head avatar~\cite{li2023animatable}, yielding a set of per-frame Gaussian parameter textures $\{\mathcal{G}_1, \dots, \mathcal{G}_{N}\}$ in dense correspondence.
Principal Component Analysis (PCA) is performed on these textures for each attribute—scale, position, opacity, and rotation—to build a linear Gaussian splat head model.
Expressions are described via low-dimensional coefficients $\mathbf{k}_i$, which define linear combinations of the attribute-specific bases $\mathbf{B}_i$:
\begin{equation}
    \mathcal{G} = \left\{ \boldsymbol{\mu}_i + \mathbf{B}_i \mathbf{k}_i \mid i \in \{\boldsymbol{\alpha}, \boldsymbol{\phi}, \boldsymbol{\sigma}, \boldsymbol{\theta} \} \right\},
\end{equation}
where the corresponding means are $\boldsymbol{\mu}_{ \{\boldsymbol{\alpha}, \boldsymbol{\phi}, \boldsymbol{\sigma}, \boldsymbol{\theta} \} }$. 
Colors are modeled as expression-independent constants. 
The bases are refined against the input images using a photometric loss.
For image-based animation, two pretrained expression estimators \cite{danvevcek2022emoca, feng2021deca} extract features from input images, and a small MLP maps these features to the coefficients $\mathbf{{k}}_i$.

\paragraph{\modelname-based GEM avatars.}
The dense correspondence of the Gaussian splat textures predicted by \modelname allows us to bypass GEM's computationally expensive mesh tracking and CNN-based head avatar training.
The predicted Gaussians are transformed into a canonical space through inverse linear blend skinning, followed by GEM's PCA decomposition, bases refinement, and MLP training steps.
Contrasting GEM, we model dynamic color changes (except for the oral cavity), perform a joint PCA decomposition across all attributes, and optimize the PCA mean $\boldsymbol{\mu}_*$.  
Please refer to \Cref{sec:supp_avatars} for more details. 

\section{Experiments}
\subsection{Datasets}

We train and evaluate \modelname on Ava-256~\cite{martinez2024codec} and NeRSemble v2~\cite{kirschstein2023nersemble}, which contain head-centric, multi-view video sequences of 256/425 subjects performing various facial expressions, captured by 80/16 cameras in a uniformly-lit studio environment.
Ava-256 provides a $360^\circ$ viewpoint coverage, whereas NeRSemble's cameras are restricted to $\pm50^\circ$ horizontally and $\pm15^\circ$ vertically. 
Ava-256 additionally includes registration meshes with hair proxy geometry for all frames.
For validation, we held out 11/6 subjects on Ava-256 and NeRSemble. 

\subsection{Novel View Synthesis}
\label{sec:nvs}

\begin{figure*}[!t]
    \centering
    \def\svgwidth{\linewidth}
    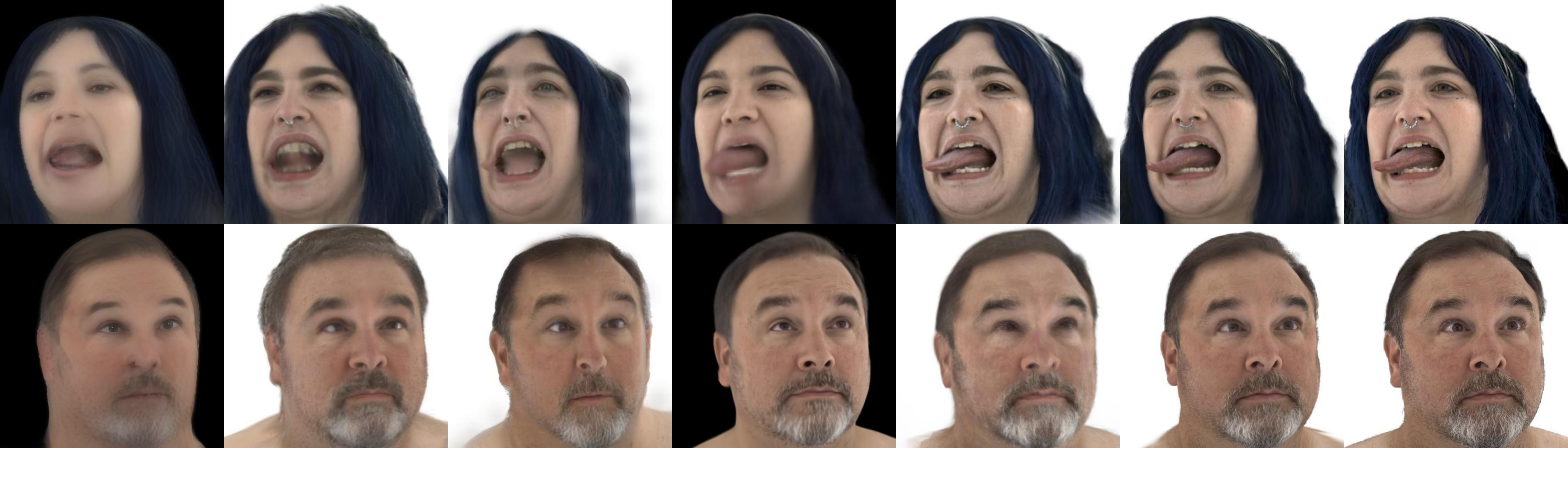
    \caption{Novel view synthesis comparison on Ava-256~\cite{martinez2024codec}. \modelname exhibits superior synthesis quality.}
    \label{fig:nvs}
\end{figure*}

\begin{table}[t]  %
\resizebox{\linewidth}{!}{%
\begin{tabular}{c|l|rrrrrr}
\toprule
Dataset & Method & LPIPS $\downarrow$ & CSIM $\uparrow$ & PSNR $\uparrow$ & SSIM $\uparrow$ & L1 $\downarrow$ & L2 $\downarrow$ \\
\midrule
\multirow{6}{*}{\rotatebox{90}{Ava-256~\cite{martinez2024codec}}}
 & GPAvatar~\cite{chu2024gpavatar} & 0.303 & 0.507 & 21.304 & 0.714 & 0.046 & 0.009 \\
 & FastAvatar~\cite{wu2025fastavatar} & 0.285 & 0.667 & 15.880 & 0.730 & 0.074 & 0.032 \\
 & LAM~\cite{he2025lam} & 0.273 & 0.678 & 15.898 & 0.734 & 0.078 & 0.035 \\
 & Avat3r~\cite{kirschstein2025avat3r} & 0.274 & 0.626 & \cellcolor{tabsecond}22.722 & 0.745 & 0.039 & \cellcolor{tabfirst}0.007 \\
 & FaceLift~\cite{lyu2025facelift} & \cellcolor{tabsecond}0.208 & \cellcolor{tabsecond}0.868 & 21.661 & \cellcolor{tabsecond}0.825 & \cellcolor{tabsecond}0.038 & 0.010 \\
 & Ours & \cellcolor{tabfirst}0.163 & \cellcolor{tabfirst}0.928 & \cellcolor{tabfirst}23.680 & \cellcolor{tabfirst}0.848 & \cellcolor{tabfirst}0.027 & \cellcolor{tabsecond}0.008 \\
\midrule
\multirow{7}{*}{\rotatebox{90}{NeRSemble~\cite{kirschstein2023nersemble}}}
 & GPAvatar~\cite{chu2024gpavatar} & 0.259 & 0.599 & \cellcolor{tabsecond}25.296 & 0.801 & 0.029 & \cellcolor{tabfirst}0.003 \\
 & FastAvatar~\cite{wu2025fastavatar} & 0.248 & 0.725 & 18.676 & 0.797 & 0.050 & 0.018 \\
 & LAM~\cite{he2025lam} & 0.254 & 0.746 & 16.996 & 0.789 & 0.062 & 0.027 \\
 & FaceLift~\cite{lyu2025facelift} & 0.200 & 0.866 & 21.524 & 0.853 & 0.040 & 0.009 \\
 & Ours (Ava-256 only) & 0.182 & 0.892 & 23.182 & 0.861 & 0.030 & 0.005 \\
 & Ours (NeRSemble only) & \cellcolor{tabsecond}0.168 & \cellcolor{tabsecond}0.927 & 24.136 & \cellcolor{tabsecond}0.870 & \cellcolor{tabsecond}0.026 & 0.005 \\
 & Ours & \cellcolor{tabfirst}0.152 & \cellcolor{tabfirst}0.944 & \cellcolor{tabfirst}25.509 & \cellcolor{tabfirst}0.884 & \cellcolor{tabfirst}0.024 & \cellcolor{tabfirst}0.003 \\
\bottomrule
\end{tabular}
}
\caption{Novel view synthesis results on Ava-256 and NeRSemble.}
\label{tab:quant_nvs}
\end{table}

We compare \modelname against single-view and multi-view 3D head reconstruction methods.
LAM~\cite{he2025lam} and FastAvatar~\cite{wu2025fastavatar} employ transformers to predict 3DMM-attached Gaussians while GPAvatar~\cite{chu2024gpavatar} generates animatable triplanes.
Avat3r~\cite{kirschstein2025avat3r} and FaceLift~\cite{lyu2025facelift} predict pixel-aligned Gaussian splats from the multi-view input images.
We provide 12 input images at $640\times512$ resolution and render a random disjoint target view.
For the single-image method LAM, the one with the smallest Euclidean distance to the target camera center is selected.

As shown in \Cref{fig:nvs}, GPAvatar's reconstructions are blurry.
FastAvatar and LAM achieve higher synthesis quality, but struggle with extreme expressions and exhibit identity shift. 
FaceLift and Avat3r produce more plausible results, yet FaceLift suffers from oversaturation artifacts and blurry textures, while Avat3r exhibits a loss of identity, especially in the mouth (top) and eye region (bottom).
Overall, \modelname reconstructs the target subjects most faithfully, even for extreme expressions such as tongue protrusion.
As CAP4D~\cite{taubner2025cap4d} only held out two Ava-256 subjects for testing, we only qualitatively compare to it on the single overlapping test subject in \Cref{fig:nvs_cap4d}.
We find that \modelname reconstructs extreme expressions better than CAP4D.

\Cref{tab:quant_nvs} quantitatively compares \modelname on 1,000 samples from Ava-256 using standard metrics such as perceptual similarity (LPIPS), Peak Signal to Noise Ratio (PSNR), Structural Similarity Metric (SSIM), pixel-wise L1 and L2 errors, and the cosine similarity between identity vectors extracted by a face recognition network~\cite{deng2019arcface} (CSIM).
We find that \modelname consistently outperforms all baselines.

\paragraph{Application to datasets w/o geometry.}
\begin{figure*}[!t]
    \centering
    \def\svgwidth{\linewidth}
    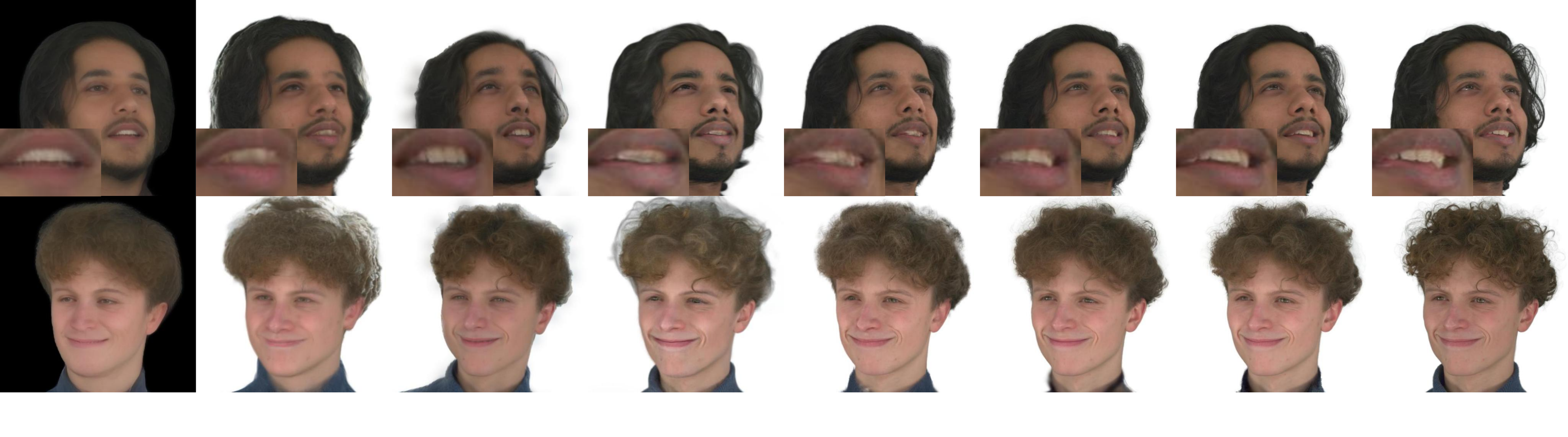
    \caption{Novel view synthesis on NeRSemble. Ours (Ava) / Ours (NeRSemble) are trained on Ava-256 and NeRSemble only, respectively.}
    \label{fig:nvs_nersemble}
\end{figure*}
We train \modelname using the 3D head registrations available in Ava-256 as supervision.
Although public datasets with larger subject counts exist~\cite{kirschstein2023nersemble, renderme360, Yang_2020_facescape}, they lack registered meshes.
This raises two questions: $i)$ What is the impact of incorporating NeRSemble during training, despite the absence of ground-truth registrations? and $ii)$ How well does \modelname generalize across datasets?
To answer these questions, \Cref{fig:nvs_nersemble} and \Cref{tab:quant_nvs} compare \modelname (trained jointly on Ava-256 and NeRSemble) against two model variants: one trained exclusively on Ava-256, and one trained exclusively on NeRSemble using pseudo-ground-truth meshes obtained via VHAP~\cite{qian2024vhap}.
Note that the NeRSemble-only variant uses the FLAME mesh topology provided by VHAP, whereas the other versions utilize the Ava-256 topology.
We evaluate these variants, alongside other baselines, on 1,000 NeRSemble samples.
Even when trained solely on Ava-256, \modelname generates plausible synthesis results on the unseen NeRSemble dataset, outperforming all baselines.
While training on NeRSemble alone improves performance, joint training on both datasets yields the best results, despite lacking geometry supervision for the NeRSemble samples.
\modelname circumvents the need for expensive head registration on NeRSemble (estimated 216k GPU-hours for 65M frames) by enabling direct training on the images.

\paragraph{Ablations.}
\begin{figure}[t]
    \centering
    \def\svgwidth{\linewidth}
    \begingroup%
  \makeatletter%
  \providecommand\color[2][]{%
    \errmessage{(Inkscape) Color is used for the text in Inkscape, but the package 'color.sty' is not loaded}%
    \renewcommand\color[2][]{}%
  }%
  \providecommand\transparent[1]{%
    \errmessage{(Inkscape) Transparency is used (non-zero) for the text in Inkscape, but the package 'transparent.sty' is not loaded}%
    \renewcommand\transparent[1]{}%
  }%
  \providecommand\rotatebox[2]{#2}%
  \newcommand*\fsize{\dimexpr\f@size pt\relax}%
  \newcommand*\lineheight[1]{\fontsize{\fsize}{#1\fsize}\selectfont}%
  \ifx\svgwidth\undefined%
    \setlength{\unitlength}{1539.1511788bp}%
    \ifx\svgscale\undefined%
      \relax%
    \else%
      \setlength{\unitlength}{\unitlength * \real{\svgscale}}%
    \fi%
  \else%
    \setlength{\unitlength}{\svgwidth}%
  \fi%
  \global\let\svgwidth\undefined%
  \global\let\svgscale\undefined%
  \makeatother%
  \small
  \begin{picture}(1,0.61152046)%
    \lineheight{1}%
    \setlength\tabcolsep{0pt}%
    \put(0,0){\includegraphics[width=\unitlength,page=1]{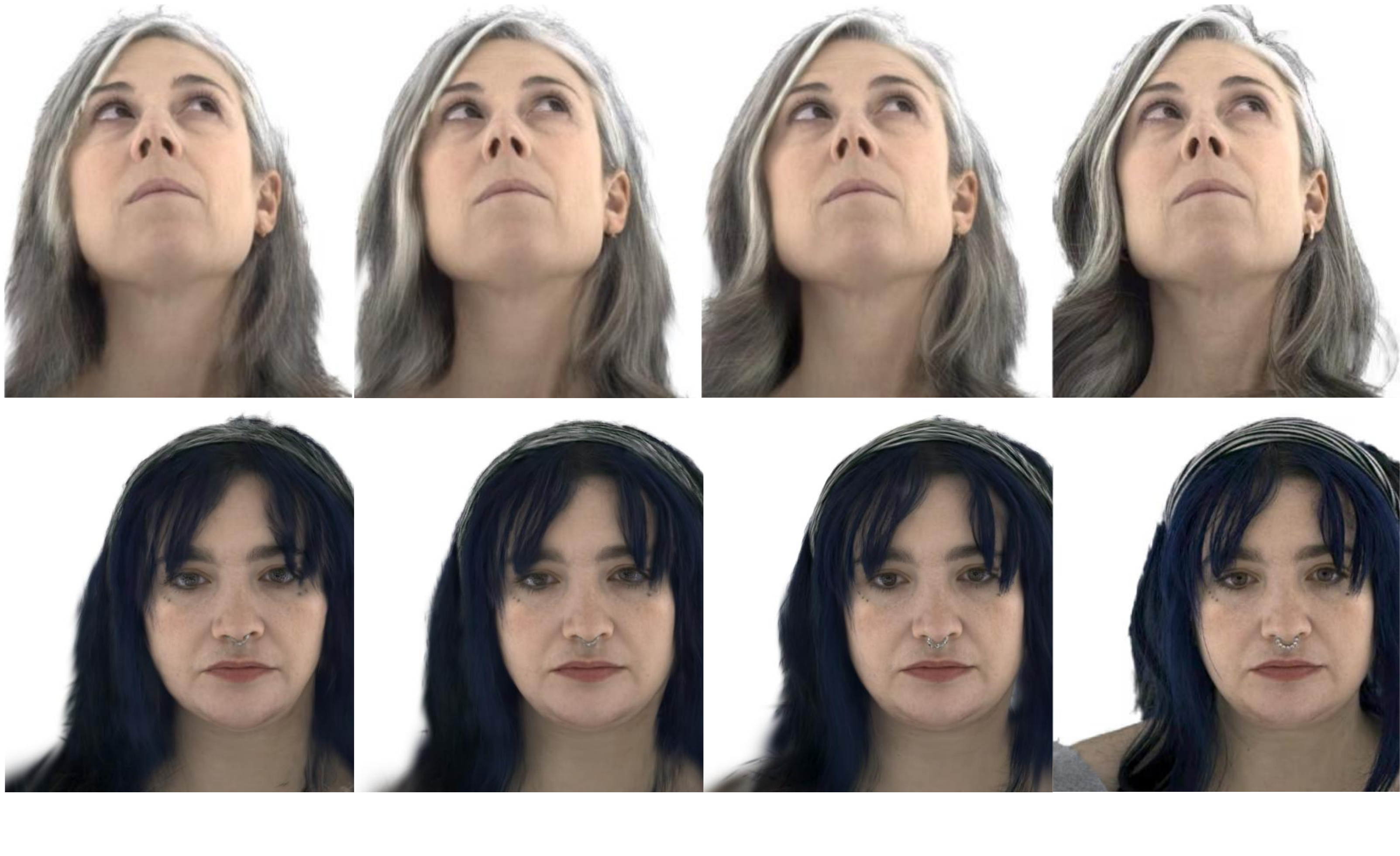}}%
    \put(0.37627963,0.00440357){\color[rgb]{0,0,0}\makebox(0,0)[t]{\lineheight{0}\smash{\begin{tabular}[t]{c}w/o Sapiens\end{tabular}}}}%
    \put(0.12679146,0.00440353){\color[rgb]{0,0,0}\makebox(0,0)[t]{\lineheight{0}\smash{\begin{tabular}[t]{c}Dense Attention\end{tabular}}}}%
    \put(0.6257678,0.00440357){\color[rgb]{0,0,0}\makebox(0,0)[t]{\lineheight{0}\smash{\begin{tabular}[t]{c}Ours\end{tabular}}}}%
    \put(0.87525593,0.00440357){\color[rgb]{0,0,0}\makebox(0,0)[t]{\lineheight{0}\smash{\begin{tabular}[t]{c}Ground Truth\end{tabular}}}}%
    \put(0,0){\includegraphics[width=\unitlength,page=2]{nvs_ablation_subset_v2.pdf}}%
  \end{picture}%
\endgroup%

    \caption{Ablation experiment on Ava-256.}
    \label{fig:nvs_ablation}
\end{figure}
\begin{table}  %
\resizebox{\linewidth}{!}{%
\begin{tabular}{l|rrrrrr}
\toprule 
 & LPIPS $\downarrow$ & CSIM $\uparrow$ & PSNR $\uparrow$ & SSIM $\uparrow$ & L1 $\downarrow$ & L2 $\downarrow$ \\
\midrule
Dense Attention & 0.221 & 0.849 & 20.364 & 0.794 & 0.041 & 0.013 \\
w/o Sapiens & 0.202 & 0.907 & 22.104 & 0.819 & 0.034 & 0.010 \\
w/o Skipconn. & 0.192 & 0.913 & \cellcolor{tabfirst}23.075 & 0.816 & 0.032 & 0.009 \\
Orig. TEMPEH & \cellcolor{tabsecond}0.190 & 0.909 & 22.775 & \cellcolor{tabsecond}0.823 & 0.033 & 0.009 \\
$H_\text{uv}=W_\text{uv}=256$ & 0.194 & 0.907 & 22.830 & 0.817 & 0.033 & 0.009 \\
$H_\text{uv}=W_\text{uv}=512$ & 0.190 & \cellcolor{tabsecond}0.914 & 22.829 & 0.822 & 0.032 & 0.009 \\
Ours & \cellcolor{tabfirst}0.187 & \cellcolor{tabfirst}0.918 & \cellcolor{tabsecond}23.032 & \cellcolor{tabfirst}0.825 & 0.032 & 0.009 \\
\bottomrule
\end{tabular}
}
\caption{Ablation experiments on Ava-256.%
}
\label{tab:ablation_quant}
\end{table}
\begin{figure}[t]
    \centering
    \def\svgwidth{\linewidth}
    \begingroup%
  \makeatletter%
  \providecommand\color[2][]{%
    \errmessage{(Inkscape) Color is used for the text in Inkscape, but the package 'color.sty' is not loaded}%
    \renewcommand\color[2][]{}%
  }%
  \providecommand\transparent[1]{%
    \errmessage{(Inkscape) Transparency is used (non-zero) for the text in Inkscape, but the package 'transparent.sty' is not loaded}%
    \renewcommand\transparent[1]{}%
  }%
  \providecommand\rotatebox[2]{#2}%
  \newcommand*\fsize{\dimexpr\f@size pt\relax}%
  \newcommand*\lineheight[1]{\fontsize{\fsize}{#1\fsize}\selectfont}%
  \ifx\svgwidth\undefined%
    \setlength{\unitlength}{1343.99990773bp}%
    \ifx\svgscale\undefined%
      \relax%
    \else%
      \setlength{\unitlength}{\unitlength * \real{\svgscale}}%
    \fi%
  \else%
    \setlength{\unitlength}{\svgwidth}%
  \fi%
  \global\let\svgwidth\undefined%
  \global\let\svgscale\undefined%
  \makeatother%
  \small
  \begin{picture}(1,0.3471)%
    \lineheight{1}%
    \setlength\tabcolsep{0pt}%
    \put(0,0){\includegraphics[width=\unitlength,page=1]{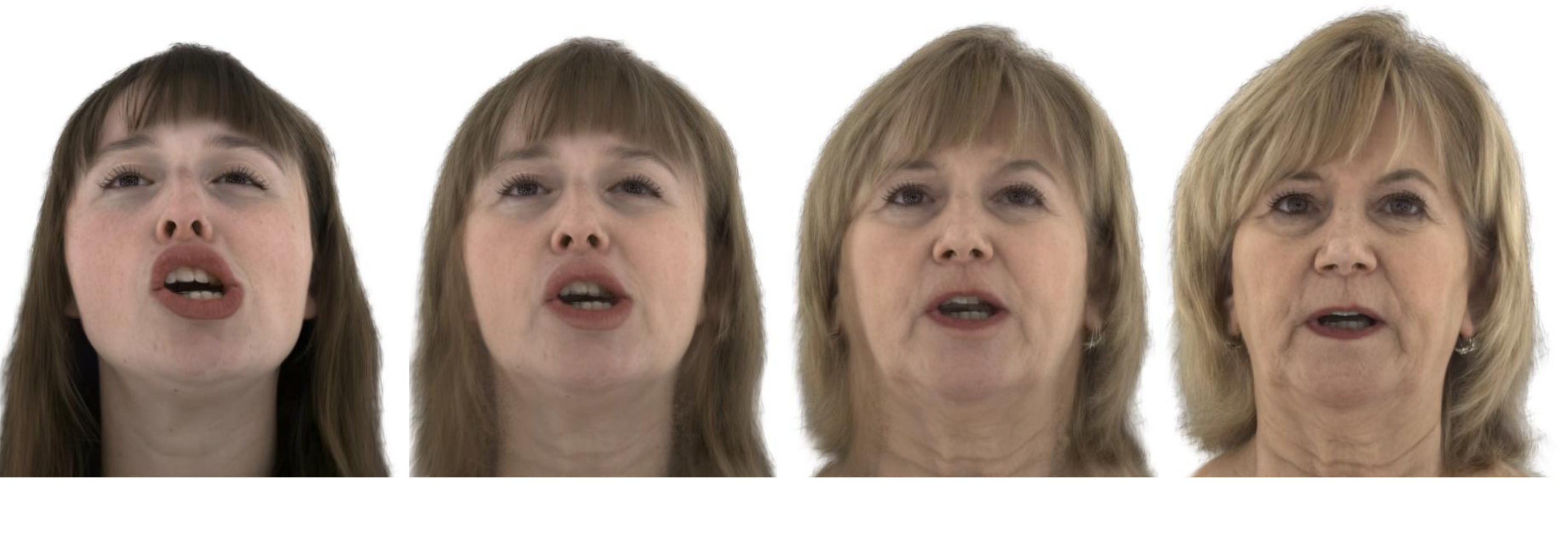}}%
    \put(0.12499988,0.00781425){\color[rgb]{0,0,0}\makebox(0,0)[t]{\lineheight{0}\smash{\begin{tabular}[t]{c}$\gamma=0.00$\end{tabular}}}}%
    \put(0.37499991,0.00781425){\color[rgb]{0,0,0}\makebox(0,0)[t]{\lineheight{0}\smash{\begin{tabular}[t]{c}$\gamma=0.33$\end{tabular}}}}%
    \put(0.62499994,0.00781425){\color[rgb]{0,0,0}\makebox(0,0)[t]{\lineheight{0}\smash{\begin{tabular}[t]{c}$\gamma=0.67$\end{tabular}}}}%
    \put(0.87499998,0.00781425){\color[rgb]{0,0,0}\makebox(0,0)[t]{\lineheight{0}\smash{\begin{tabular}[t]{c}$\gamma=1.00$\end{tabular}}}}%
  \end{picture}%
\endgroup%

    \caption{Interpolation between Gaussian textures by factor $\gamma$.}  %
    \label{fig:interpolation}
\end{figure}
\begin{figure}[t]
    \centering
    \def\svgwidth{\linewidth}
    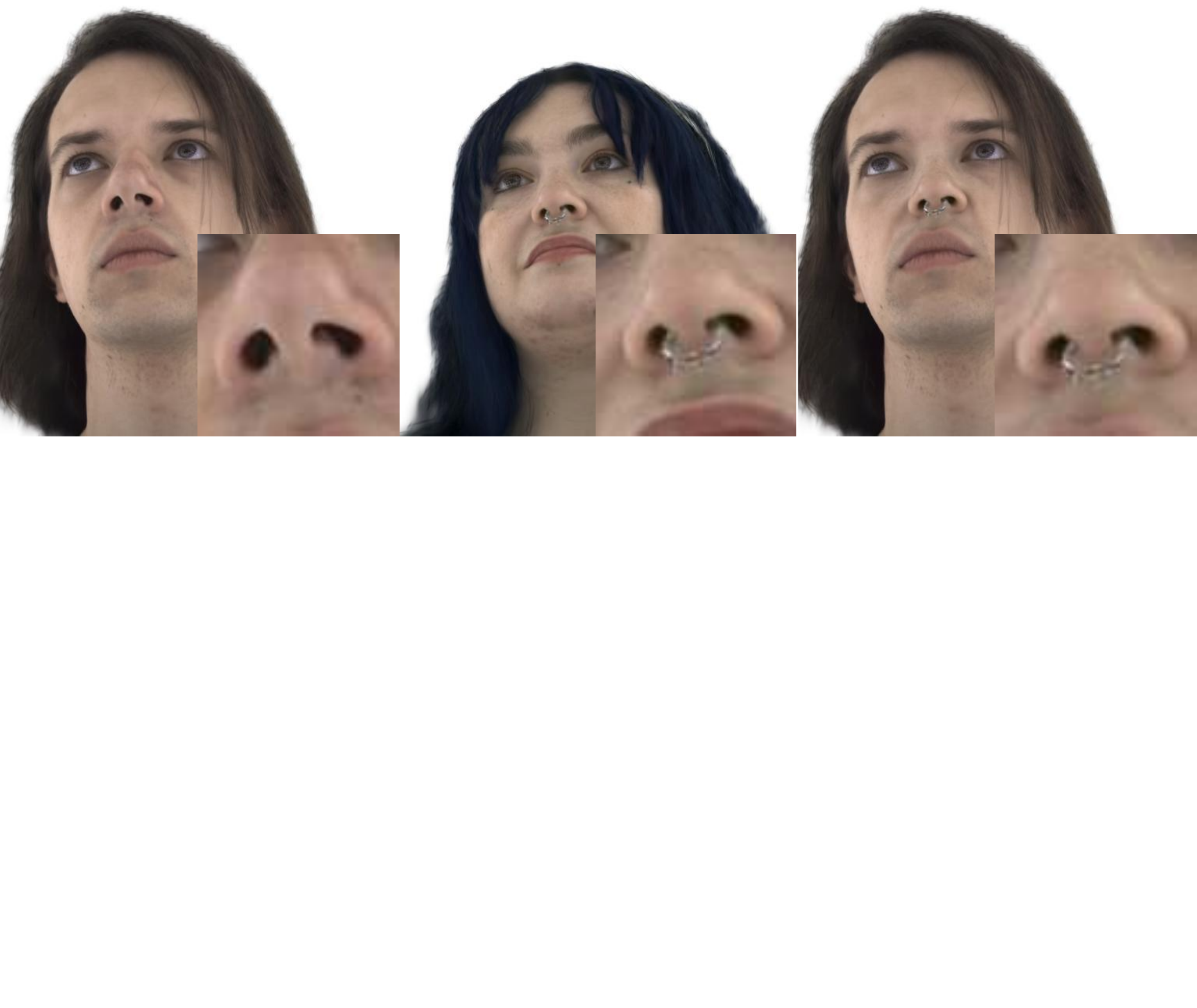
    \caption{
    (Top) Semantic editing: Replacing a source’s nose with that of a target.
    (Bottom) Expression transfer: Jaw articulation and mouth interior are transferred from a target to a source identity.}
    \label{fig:editing_expr_transfer}
\end{figure}
\Cref{tab:ablation_quant} evaluates the impact of individual model design choices and \Cref{fig:nvs_ablation} provides visualizations for the most influential components.
All models are trained for 300k iterations on Ava-256.
We find that the registration-guided attention yields the biggest performance gain; performing dense attention across all UV and image tokens ('Dense Attention') instead gives the worst scores.
Qualitatively, we observe less fine detail in face accessories and worse hair textures with the dense attention approach.
The pretrained Sapiens~\cite{khirodkar2024sapiens} feature extractor has the second-biggest impact.
Disabling it ('w/o Sapiens') worsens generalization and reduces the synthesis quality.
Further, the reconstruction quality improves as we increase the UV texture resolution.
Additionally, both removing the skip connections ('w/o Skipconn.') and using the original global TEMPEH model for geometry initialization instead of our adapted version ('Orig. TEMPEH'), produce slightly worse reconstructions.
\Cref{sec:supp_ablations} provides additional ablations on the number of input images and the number of image tokens $k_{\mathcal{T},\text{img}}$ to attend to in the registration-guided attention blocks.
We find that \modelname produces plausible reconstructions from only four input images and that smaller $k_{\mathcal{T},\text{img}}$ improve performance.

\subsection{Interpolation, Editing \& Expression Transfer}
\label{sec:editing}

The dense semantic correspondence predicted by \modelname enables diverse applications, including cross-identity and cross-expression interpolation, part-based editing, and expression transfer.
\Cref{fig:interpolation} shows that simple interpolation between two subjects with different expressions gives smooth and plausible intermediate results.
\Cref{fig:editing_expr_transfer} (top) demonstrates semantic editing, where Gaussian splat attributes in the nose region of a source identity are replaced with the ones from a target subject, resulting in a seamless blend.
Additionally, \Cref{fig:editing_expr_transfer} (bottom) illustrates an arithmetic expression transfer approach, where the residual of Gaussian maps for an expressive and a neutral frame of a target subject is added to the neutral reconstruction of a source identity, resulting in a plausible expression transfer.
Please refer to \Cref{fig:editing_b} and \Cref{fig:expr_transfer_b} for more results.

\subsection{Geometry Reconstruction}
\label{sec:mesh_extraction}
\begin{figure}[t]
    \centering
    \def\svgwidth{\linewidth}
    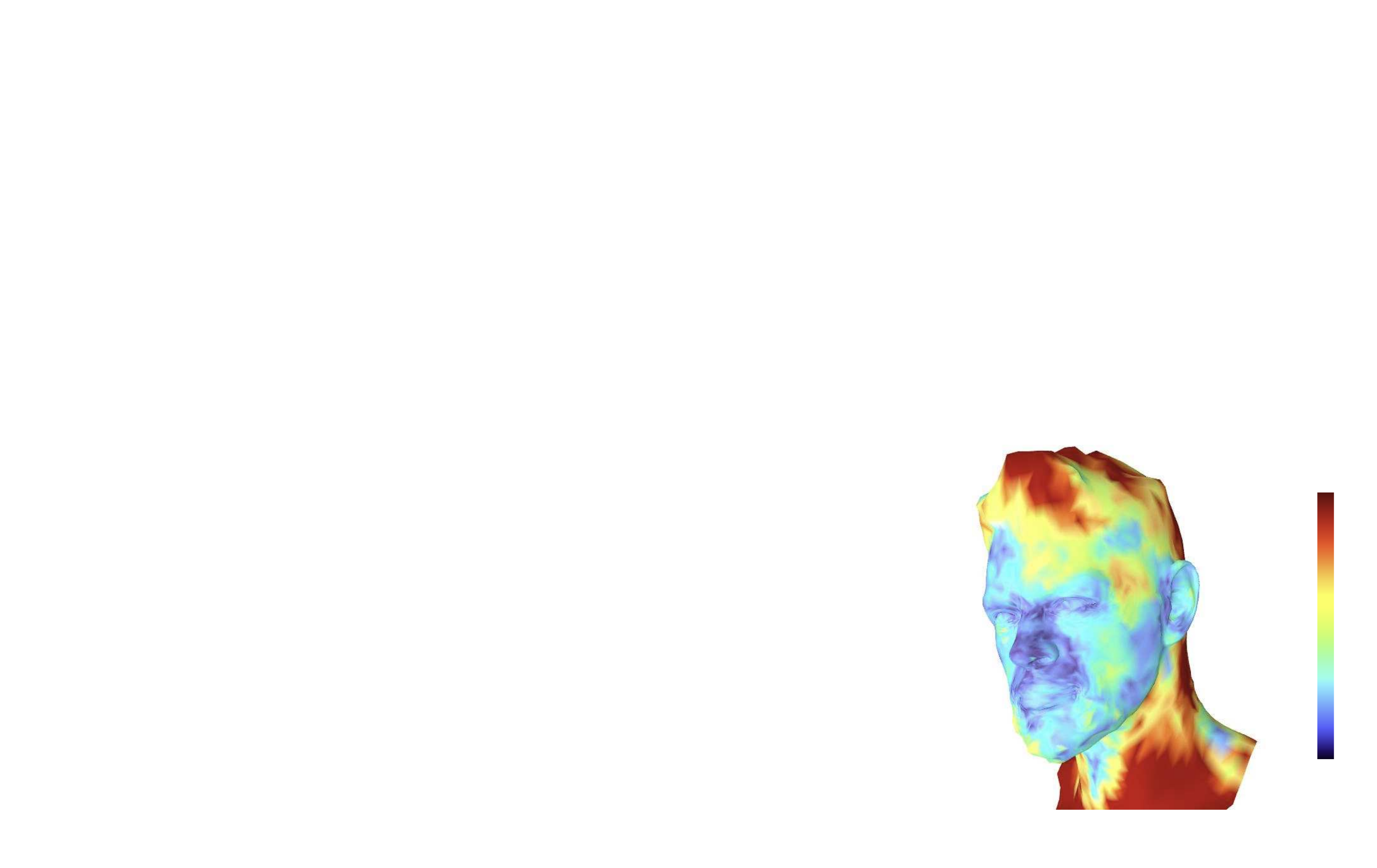
    \caption{Geometry reconstruction comparison. The heatmaps visualize the Euclidean distance between the predicted and ground truth vertices. 'TEMPEH Glob.$^*$' is the adapted version of TEMPEH's global stage that we use for the coarse mesh registration. }
    \label{fig:geometry_recon}
\end{figure}

\begin{table}  %
\resizebox{\linewidth}{!}{%
\begin{tabular}{l|rr|rr|rr|rr}
\toprule 
 &\multicolumn{2}{c|}{Full}&\multicolumn{2}{c|}{Face}&\multicolumn{2}{c|}{Mouth}&\multicolumn{2}{c}{Eyes}\\
 &P2P $\downarrow$ & P2S $\downarrow$&P2P $\downarrow$ & P2S $\downarrow$&P2P $\downarrow$ & P2S $\downarrow$&P2P $\downarrow$ & P2S $\downarrow$\\
\midrule
TEMPEH Glob.$^*$ & \cellcolor{tabsecond}7.34 & 2.76 & \cellcolor{tabsecond}3.84 & 1.50 & \cellcolor{tabsecond}2.69 & 1.15 & \cellcolor{tabsecond}2.27 & 1.09 \\
TEMPEH~\cite{bolkart2023tempeh} & 7.84 & \cellcolor{tabsecond}2.18 & 4.04 & \cellcolor{tabsecond}0.97 & 2.99 & \cellcolor{tabfirst}0.69 & 2.66 & \cellcolor{tabsecond}0.62 \\
Ours & \cellcolor{tabfirst}6.69 & \cellcolor{tabfirst}2.10 & \cellcolor{tabfirst}3.18 & \cellcolor{tabfirst}0.88 & \cellcolor{tabfirst}2.14 & \cellcolor{tabsecond}0.72 & \cellcolor{tabfirst}1.54 & \cellcolor{tabfirst}0.61 \\
\bottomrule
\end{tabular}
}
\caption{Quantitative geometry reconstruction comparison on Ava-256. Scores are reported in mm. P2P: Point-to-Point distance. P2S: Point-to-Surface distance. 'TEMPEH Glob.$^*$' is the adapted version of TEMPEH's global stage that we use for the coarse mesh registration. 'Face' is the facial area w/o mouth, eyes, and ears.}
\label{tab:quant_geom}
\end{table}

\begin{figure*}[!t]
    \centering
    \def\svgwidth{\linewidth}
    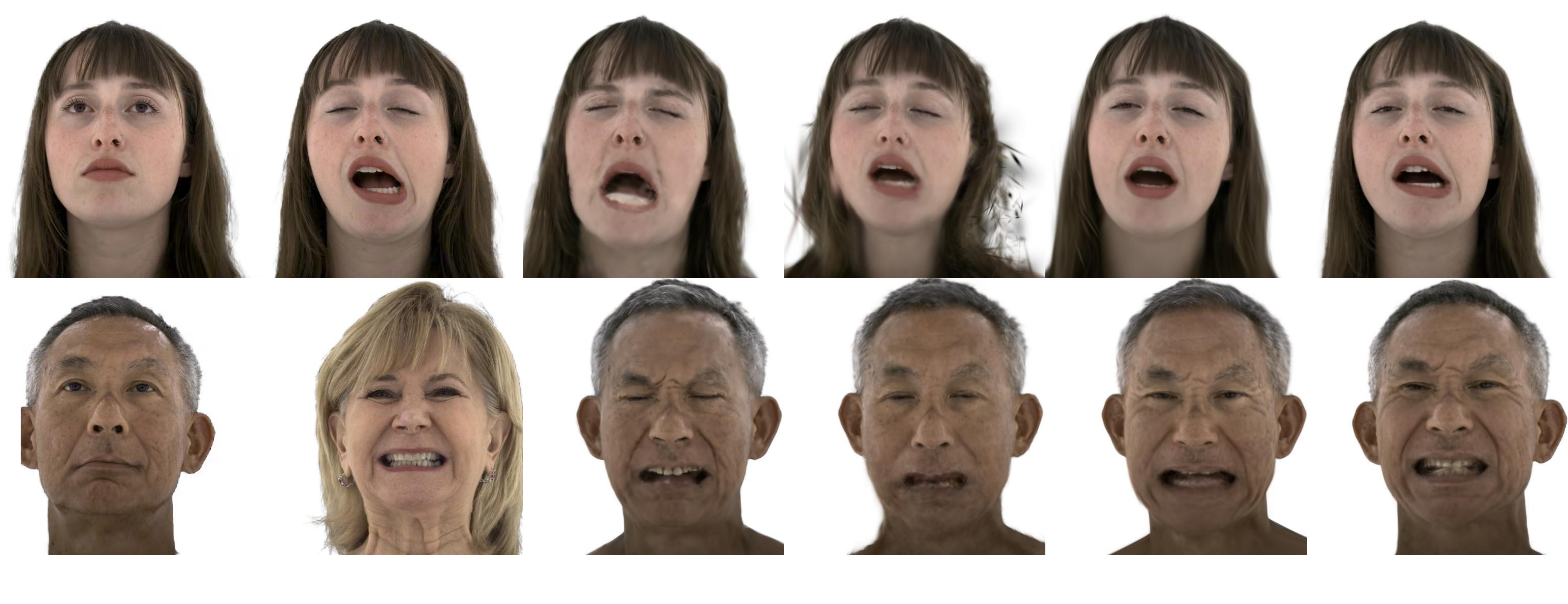
    \caption{Qualitative comparison for image-based self-reenactment (top) and cross-reenactment (bottom)  of the subject-specific avatars.}
    \label{fig:avatars_qual}
\end{figure*}

\modelname predicts texture maps of Gaussian splat attributes, including a map of 3D positions.
Given vertex UV coordinates, these location textures can be converted into a mesh, making \modelname an inference-based head tracker. 
We compare our geometry reconstruction against TEMPEH~\cite{bolkart2023tempeh}, a state-of-the-art tracker that estimates registered meshes from multi-view images in a single inference step.
For a fair comparison, we retrain TEMPEH on the Ava-256 dataset using the provided mesh topology.
We also include the adapted TEMPEH global stage, which generates the coarse mesh registrations $\mathcal{M}$ used as input to our method.
\Cref{tab:quant_geom} reports the Euclidean vertex-to-vertex distance (P2P) and the Point-to-Surface (P2S) distance between the predicted and ground-truth meshes on 1,000 Ava-256 samples, while a qualitative comparison can be found in \Cref{fig:geometry_recon}.
We observe that our method outperforms both baselines on both metrics when averaged over the entire head.  
In the mouth and eye regions, our method performs comparably to or slightly worse than TEMPEH.
This is expected, as the ground truth meshes approximate the oral cavity and eyes with planar faces, while \modelname deviates from this coarse approximation to reconstruct the appearance necessary for a photorealistic rendering. 

\subsection{Subject-Specific Head Avatars}

\begin{table}  %
\resizebox{\linewidth}{!}{%
\begin{tabular}{l|r|rrr|rr}
\toprule 
& Reconstr. & \multicolumn{3}{c|}{Self-Reenactment} & \multicolumn{2}{c}{Cross-Reenactment}\\
& Time $\downarrow$ &LPIPS $\downarrow$ & SSIM $\uparrow$ & PSNR $\uparrow$ & CSIM $\uparrow$ & EmoL1 $\downarrow$ \\
\midrule
GA~\cite{qian2024gaussianavatars} & 15.5h & 0.233 & 0.755 & 21.248 & 0.629 & 10.618 \\
RGBAvatar~\cite{li2025rgbavatar} & \cellcolor{tabsecond}11.4h & 0.233 & \cellcolor{tabsecond}0.782 & \cellcolor{tabsecond}22.185 & 0.657 & 10.333 \\
GEM~\cite{zielonka2025gem} & 45.3h & \cellcolor{tabsecond}0.214 & 0.778 & 21.761 & \cellcolor{tabsecond}0.800 & \cellcolor{tabsecond}9.866 \\
Ours & \cellcolor{tabfirst}4.6h & \cellcolor{tabfirst}0.174 & \cellcolor{tabfirst}0.809 & \cellcolor{tabfirst}24.122 & \cellcolor{tabfirst}0.813 & \cellcolor{tabfirst}9.837 \\
\bottomrule
\end{tabular}
}
\caption{Quantitative comparison of subject-specific head avatars.}
\label{tab:avatars_quant}
\end{table}

We compare \modelname-based GEM avatars (\Cref{sec:avatar_creation}) with GEM~\cite{zielonka2025gem}, GaussianAvatars (GA)~\cite{qian2024gaussianavatars}, and RGBAvatar~\cite{li2025rgbavatar}.
All methods are optimized per-subject on Ava-256 multi-view videos, captured by 12 cameras with an average sequence length of 3,300 frames.
We evaluate the self- and cross-reenactment performances for five subjects on held-out sequences in \Cref{tab:avatars_quant} and \Cref{fig:avatars_qual}.
For image-based animation, we drive GaussianAvatars and RGBAvatar with EMOCA's~\cite{danvevcek2022emoca} parameters, which is also used as a pretrained feature extractor to drive GEM and our \modelname-based GEM avatar.
Please refer to \Cref{sec:supp_avatar_baselines} for more details. 
We measure standard image metrics for self-reenactment. 
For cross-reenactment, we evaluate the cosine similarity between features extracted with a pretrained face recognition network \cite{deng2019arcface} (CSIM), and follow GEM \cite{zielonka2025gem} in evaluating the L1 distance between features of the emotion recognition network EmoNet \cite{abdul2017emonet} (EmoL1). 

We find that our method consistently outperforms all baselines in the self-reenactment scenario and performs slightly better or on par with GEM for cross-reenactment.
As we adapt GEM's image-based encoder, a similar cross-reenactment performance is plausible.
RGBAvatar exhibits noticeable artifacts for subjects with long hair and produces blurry results for unseen expressions.  
GaussianAvatars generally handles unseen expressions more robustly, but is inferior to GEM and \modelname in terms of synthesis fidelity.

Our method not only improves synthesis quality but also drastically cuts reconstruction time.
We achieve a $10\times$ reduction relative to GEM (the baseline with the closest synthesis quality) and are $2.5\times$ faster than the fastest baseline, RGBAvatar, which exhibits inferior visual fidelity.
Please refer to \Cref{sec:supp_avatar_timing} and \Cref{sec:supp_avatar_ablation} for detailed timings and ablation experiments on the changes compared to GEM.

\section{Discussion}
\begin{figure}[t]
    \centering
    \includegraphics[width=\linewidth]{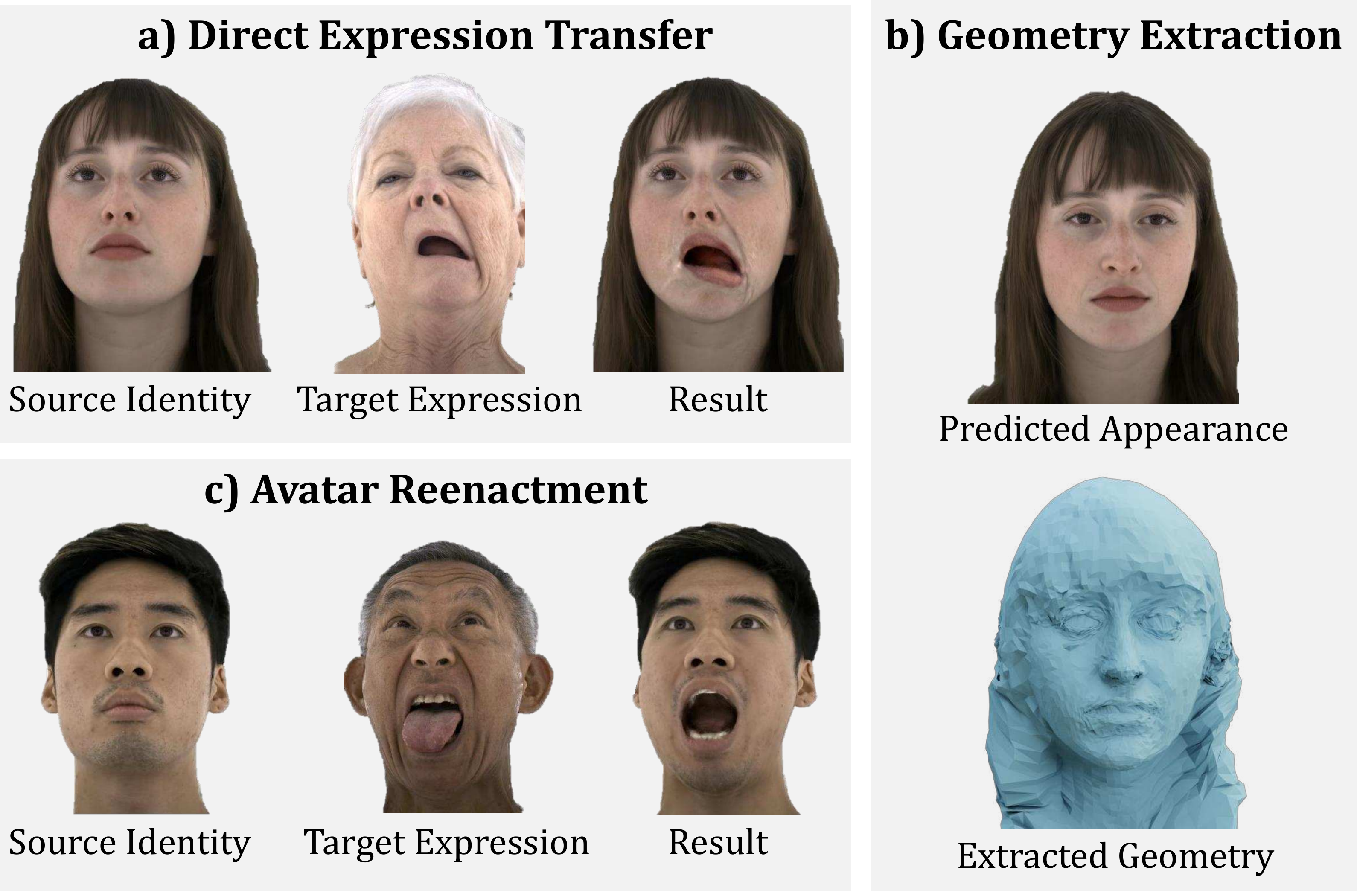}
    \caption{
    Limitations.
    a) Direct expression transfer (\Cref{sec:editing}) may incur identity leakage for dissimilar identities. %
    b) The meshes extracted from the predicted Gaussian 3D location textures (\Cref{sec:mesh_extraction}) can exhibit self-intersecting surfaces.
    c) The subject-specific head avatars (\Cref{sec:avatar_creation}) are bound to interpolations of the training expressions and do not track eye movement.
    }
    \label{fig:limitations}
\end{figure}
\paragraph{Registration dependency.}
\modelname uses Ava-256 mesh registrations as training supervision. 
While \Cref{sec:nvs} shows cross-dataset training with mesh supervision on only one dataset, training entirely without it is subject to future work.
This could potentially be achieved by training \modelname's geometry part solely on synthetic data.

\paragraph{Expression transfer.}
The direct transfer of extreme expressions between \modelname's predicted Gaussian textures sometimes results in appearance leakage (\Cref{fig:limitations}a).
A less simplistic, learning-based method for expression transfer would likely be a more suitable choice.

\paragraph{Geometry reconstruction.}
Converting the predicted Gaussian 3D location textures into registered meshes occasionally results in self-intersecting surfaces (\Cref{sec:mesh_extraction}).
While providing perfectly smooth registration meshes is not a primary goal of \modelname, adding additional surface regularization during training could mitigate such artifacts.

\paragraph{Avatar quality.}
By following GEM's \cite{zielonka2025gem} avatar formulation, the resulting avatars inherit its limitations of only generating expressions similar to those in the training data.
Moving forward, an interesting direction is to leverage $\text{\modelname}$'s cross-subject and cross-expression correspondence by learning a prior across identities and expressions, and then adapting it for per-subject avatars to achieve stronger generalization to unseen expressions.

\section{Conclusion}
We have presented \modelname, a framework that unifies geometry registration and radiance field reconstruction by predicting Gaussian splats in dense semantic correspondence.
Central to this approach is our novel registration-guided attention mechanism, which enhances both computational efficiency and synthesis quality.
By bypassing the bottleneck of traditional optimization-based tracking, \modelname reduces the total avatar creation time from 45 hours to just 4.6 hours, a $\mathit{10\times}$ speedup over the qualitatively closest baseline, while simultaneously improving visual fidelity.
Beyond efficiency, the learned intra- and inter-subject correspondence unlocks diverse applications, including semantic editing, robust geometry registration, and identity interpolation.
Overall, \modelname represents a clear step toward scalable, high-fidelity avatar creation and the structured Gaussian registration representation provides a solid basis for future work on learning priors across different subjects.

\section{Acknowledgements}

We thank W. Zielonka for data preparation and GEM training support, and B.~B.~Bilecen, B. Thambiraja, A.-L. Schweikert, S. Kocour, P.-W. Grassal, X. Lyu, J. Thies, and F. Rajič for proofreading.
MP was partially funded by the Max Planck ETH Center for Learning Systems (CLS).
Computations were performed on the MPI-IS Tübingen cluster. 
The \modelname icon is from \href{https://www.flaticon.com/free-icons/ignite}{flaticon}.

{
    \small
    \bibliographystyle{ieeenat_fullname}
    \bibliography{main.bib}
}

\appendix
\clearpage
\setcounter{page}{1}
\twocolumn[{%
    \renewcommand\twocolumn[1][]{#1}%
    \maketitlesupplementary
    \begin{center}
        \centering
        \captionsetup{type=figure}
        \def\svgwidth{\linewidth}
        \begingroup%
  \makeatletter%
  \providecommand\color[2][]{%
    \errmessage{(Inkscape) Color is used for the text in Inkscape, but the package 'color.sty' is not loaded}%
    \renewcommand\color[2][]{}%
  }%
  \providecommand\transparent[1]{%
    \errmessage{(Inkscape) Transparency is used (non-zero) for the text in Inkscape, but the package 'transparent.sty' is not loaded}%
    \renewcommand\transparent[1]{}%
  }%
  \providecommand\rotatebox[2]{#2}%
  \newcommand*\fsize{\dimexpr\f@size pt\relax}%
  \newcommand*\lineheight[1]{\fontsize{\fsize}{#1\fsize}\selectfont}%
  \ifx\svgwidth\undefined%
    \setlength{\unitlength}{3203.56345349bp}%
    \ifx\svgscale\undefined%
      \relax%
    \else%
      \setlength{\unitlength}{\unitlength * \real{\svgscale}}%
    \fi%
  \else%
    \setlength{\unitlength}{\svgwidth}%
  \fi%
  \global\let\svgwidth\undefined%
  \global\let\svgscale\undefined%
  \makeatother%
  \begin{picture}(1,0.395261)%
    \lineheight{1}%
    \setlength\tabcolsep{0pt}%
    \put(0,0){\includegraphics[width=\unitlength,page=1]{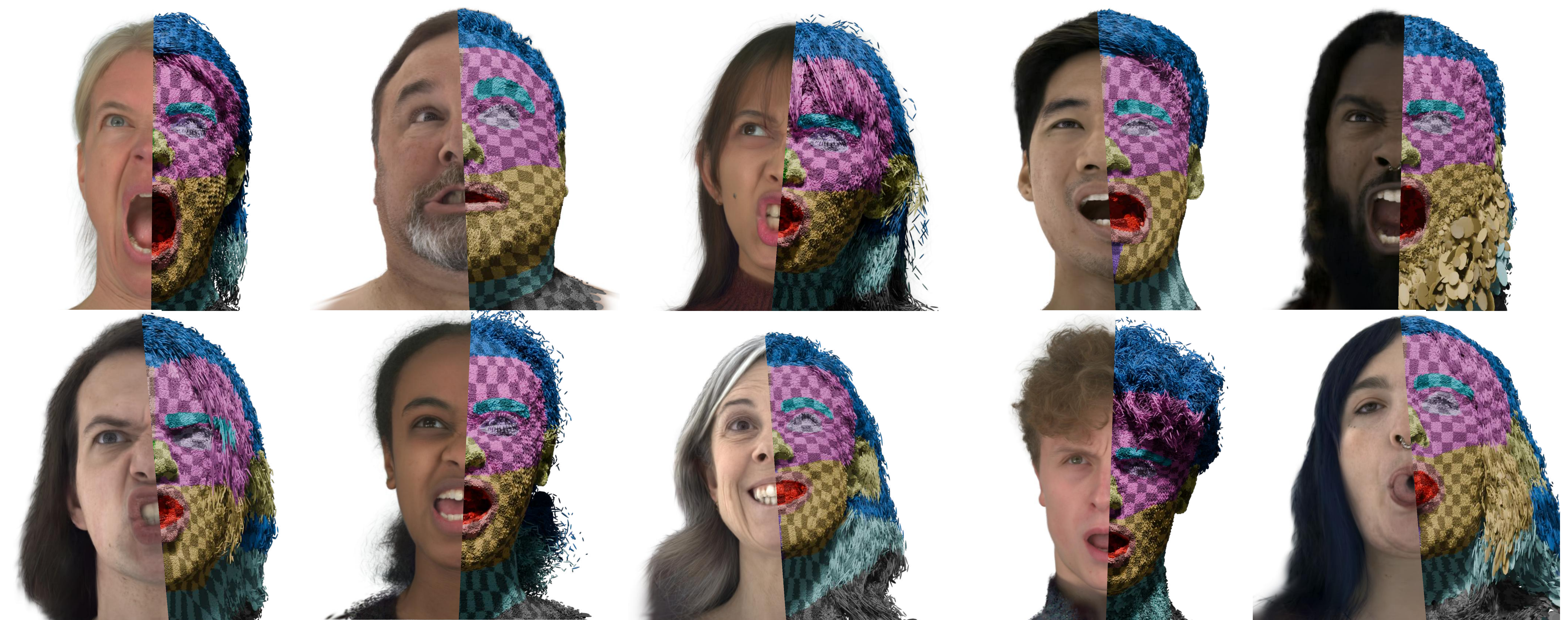}}%
  \end{picture}%
\endgroup%

        \caption{\modelname predictions overlaid with texturized Gaussian splat renderings. }
        \label{fig:suppteaser}
    \end{center}%
}]
\begin{figure*}[!t]
    \centering
\end{figure*}
\paragraph{Contents.}
This supplementary material provides additional results, implementation details, and analyses that support the claims presented in the main paper. 
\Cref{sec:supp_datasets} gives more information on the datasets used for training and evaluation, followed by implementation details in \Cref{sec:supp_implementation}.
\Cref{sec:supp_nvs} provides more novel view synthesis experiments. 
\Cref{sec:itw} and \Cref{sec:single-image} present reconstructions for in-the-wild and single-input-image scenarios respectively. 
\Cref{sec:supp_editing} shows additional results for the interpolation, semantic editing, and expression transfer on Gaussian splat textures.
\Cref{sec:quant_correspondence_eval} supplements a quantitative evaluation of the correspondences predicted by \modelname.
\Cref{sec:supp_avatars} adds further comparisons of the subject-specific head avatars created from \modelname's predictions. 
We close with ethical considerations in \Cref{sec:supp_ethics}.

\section{Datasets}
\label{sec:supp_datasets}
\paragraph{Ava-256.}
The Ava-256 dataset~\cite{martinez2024codec} provides multi-view video captures from 80 cameras of 256 subjects performing a wide range of expressions.
Ava-256 comes with ground truth mesh registrations of the head, including hair proxy geometry. 
We sample the captured videos at a framerate of 0.75 fps and select cameras that are evenly distributed over the frontal hemisphere in a range of $\pm40^\circ$ horizontally and $[-15^\circ, +36^\circ]$ vertically.
To facilitate generalization to other datasets with different backgrounds, we remove the image background using the provided alpha masks. 
We adopt Avat3r's~\cite{kirschstein2025avat3r} train-validation split and use 244 identities for training and 11 for validation. One subject was removed from Avat3r's validation set due to faulty ground truth segmentation masks. 
We sample 123,000 frames for training and 1,000 for validation.

\paragraph{NeRSemble.}
The NeRSemble v2 dataset~\cite{kirschstein2023nersemble} provides multi-view video captures from 16 cameras of 425 subjects, which are split into 419 training and 6 validation subjects. 
We select cameras 
\begin{verbatim}
221501007, 222200045, 
222200049, 222200043, 
222200047, 222200038, 
220700191, 222200041, 
222200046, 222200040, 
222200042, 222200044
\end{verbatim}
for training and leave the rest for validation.
As with Ava-256, we sample 123,000 frames for training and 1,000 for validation.
Further, we use a pretrained matting model~\cite{BGMv2} to whiten the background. 
Pseudo ground truth mesh registrations are obtained with an optimization-based head tracker~\cite{qian2024vhap}.

\section{Implementation Details}
\label{sec:supp_implementation}
\begin{table*}[]
\centering
\resizebox{0.8\linewidth}{!}{%
\begin{tabular}{l|l|l|l}
\toprule
Iteration   & Stage Name                                                       & Training Datasets                                                 & Loss Weights                                                                                                                                                         \\ \midrule
0 - 100k    & Geometry Only                                                    & Ava-256~\cite{martinez2024codec}                                                           & \begin{tabular}[c]{@{}l@{}}$w_\text{geometry}=1\times 10^{-3}$\\ $w_{\text{reg}}=1\times 10^{-3}$\\ $w_\text{lpips} = 0$\\ $w_\text{L1} = 0$\\ $w_\text{SSIM}= 0$\end{tabular}                                                            \\\midrule
100k - 400k & \begin{tabular}[c]{@{}l@{}}Geometry \\ \& Apparance\end{tabular} & Ava-256~\cite{martinez2024codec}                                                           & \begin{tabular}[c]{@{}l@{}}$w_\text{geometry}=1\times 10^{-3}$\\ $w_\text{reg}=1\times 10^{-3}$\\ $w_\text{L1} = 0.8$\\ $w_\text{SSIM}= 0.2$\\ $w_\text{LPIPS} = 0$ until 150k, then linearly increasing to $1.0$ until 200k\end{tabular} \\\midrule
400k - 860k & Mixed Training                                                   & \begin{tabular}[c]{@{}l@{}}Ava-256~\cite{martinez2024codec} \\ \& NeRSemble~\cite{kirschstein2023nersemble}\end{tabular} & \begin{tabular}[c]{@{}l@{}}$w_\text{geometry}=(\text{Ava-256: } 1\times 10^{-3}; \text{NeRSemble: } 0)$\\ $w_\text{reg}=1\times 10^{-3}$\\ $w_\text{L1} = 0.8$\\ $w_\text{SSIM}= 0.2$\\ $w_\text{LPIPS} = 1$\end{tabular}\\\bottomrule
\end{tabular}
}
\caption{\modelname loss weights in different training stages.}
\label{tab:lossweights}
\end{table*}

\begin{table}
\resizebox{\linewidth}{!}{%
\begin{tabular}{l|l}
\toprule 
Parameter & Value \\
\midrule
UV texture resolution $H_\text{uv}\times W_\text{uv}$ & $1024\times1024$ \\
UV token patch size $p_\text{uv}$& $16$ \\
Image token patch size $p_\text{img}$& $8$ \\
Token dimension $d$ & $512$ \\
Number of input images $V$& $12$ \\
Image resolution $H_\text{img}\times W_\text{img}$& $640\times512$ \\
Number of registration-guided attention blocks & $6$ \\
Registration-guided attention image token count $k_{\mathcal{T},\text{img}}$ & $100$ \\
Number of grouped attention blocks & $6$ \\
Gaussian scale regularization target & $5\times10^{-4}$ \\
Gaussian opacity regularization target & $0.7$\\
Learnable UV token positional embedding dimension & $512$ \\
\bottomrule
\end{tabular}
}
\caption{MATCH hyperparameters.}
\label{tab:hyperparameters}
\end{table}

\Cref{tab:hyperparameters} provides a list of the most important hyperparameter values, and \Cref{tab:lossweights} presents the loss weights used in the different training stages. 
We use the AdamW optimizer~\cite{loshchilov2017decoupled} with an initial learning rate of $4e-5$, a weight decay of $0.05$, and a cosine learning rate scheduler that decreases the learning rate to 0 within 1M steps after a 1,000-step linear warm-up phase. 
For calculating the Sapiens feature maps of the input images, we assemble them into grids of $2\times 2$ before feeding them into the feature extractor to save computation time.

\section{Novel View Synthesis}
\label{sec:supp_nvs}
\subsection{Detailed Baseline Description}
We compare our model against the baselines GPAvatar~\cite{chu2024gpavatar}, Fastavatar~\cite{wu2025fastavatar}, LAM~\cite{he2025lam}, Avat3r~\cite{kirschstein2025avat3r}, FaceLift~\cite{lyu2025facelift}, and CAP4D~\cite{taubner2025cap4d}.

LAM~\cite{he2025lam} predicts Gaussian splats for each vertex of a subsampled 3DMM from a single input image using a transformer backbone. These can be directly driven using 3DMM parameters.
Fastavatar~\cite{wu2025fastavatar} builds on this approach and enables the aggregation of information extracted from several input images. 
GPAvatar~\cite{chu2024gpavatar} follows a different approach and reconstructs 3D head avatars from one or several input images using a triplane representation that can be animated with point-based expression fields. 

FaceLift~\cite{lyu2025facelift} trains a Gaussian Splatting Large Reconstruction Model (GS-LRM)~\cite{zhang2024gs} on synthetic data to predict pixel-aligned Gaussian splats from several input images. 
This GS-LRM is used to lift predictions of a diffusion model, which infers multi-view images from a single reference, into a 3D Gaussian splatting representation. 
In our comparison, we solely focus on the GS-LRM model of FaceLift, which receives the ground truth multi-view images as input. 
Avat3r~\cite{kirschstein2025avat3r} similarly regresses pixel-aligned Gaussian splats, yet they can be directly animated into new expressions through cross-attention to latent expression codes. 
Note that these latent expression codes are constructed from high-quality mesh registrations and texture reprojections that are obtained with closed-source software, making Avat3r inapplicable to datasets other than Ava-256. 
The Gaussian splats predicted by FaceLift and Avat3r are pixel-aligned and do not exhibit any semantic correspondence across frames or subjects. 

CAP4D~\cite{taubner2025cap4d} uses a 3DMM-conditioned multi-view diffusion model to generate images with novel pose and expression, given one or several input images. 
In contrast to the other baselines, which infer 3D representations, it predicts 2D images that are not truly 3D-consistent.  
CAP4D only held out two subjects of the Ava-256 dataset for validation, only one of which intersects with the validation subjects from Avat3r and our method. 
As a consequence, we only perform a qualitative comparison with CAP4D on this one subject (see \Cref{fig:nvs_cap4d}). 

Since different methods predict different crops of the face, we evaluate the methods on the maximum square crop that fits into the intersection of all bounding boxes, resized to a resolution of $512\times512$, and mask out the torso and shoulders using a pretrained segmentation network \cite{khirodkar2024sapiens}. 

\subsection{Further qualitative comparisons}
\begin{figure}[t]
    \centering
    \def\svgwidth{\linewidth}
    \begingroup%
  \makeatletter%
  \providecommand\color[2][]{%
    \errmessage{(Inkscape) Color is used for the text in Inkscape, but the package 'color.sty' is not loaded}%
    \renewcommand\color[2][]{}%
  }%
  \providecommand\transparent[1]{%
    \errmessage{(Inkscape) Transparency is used (non-zero) for the text in Inkscape, but the package 'transparent.sty' is not loaded}%
    \renewcommand\transparent[1]{}%
  }%
  \providecommand\rotatebox[2]{#2}%
  \newcommand*\fsize{\dimexpr\f@size pt\relax}%
  \newcommand*\lineheight[1]{\fontsize{\fsize}{#1\fsize}\selectfont}%
  \ifx\svgwidth\undefined%
    \setlength{\unitlength}{1151.9999827bp}%
    \ifx\svgscale\undefined%
      \relax%
    \else%
      \setlength{\unitlength}{\unitlength * \real{\svgscale}}%
    \fi%
  \else%
    \setlength{\unitlength}{\svgwidth}%
  \fi%
  \global\let\svgwidth\undefined%
  \global\let\svgscale\undefined%
  \makeatother
\small%
  \begin{picture}(1,1.0540472)%
    \lineheight{1}%
    \setlength\tabcolsep{0pt}%
    \put(0.16482846,0.00196382){\color[rgb]{0,0,0}\makebox(0,0)[t]{\lineheight{0}\smash{\begin{tabular}[t]{c}CAP4D~\cite{taubner2025cap4d}\end{tabular}}}}%
    \put(0.50000002,0.00196382){\color[rgb]{0,0,0}\makebox(0,0)[t]{\lineheight{0}\smash{\begin{tabular}[t]{c}Ours\end{tabular}}}}%
    \put(0.83149518,0.00196382){\color[rgb]{0,0,0}\makebox(0,0)[t]{\lineheight{0}\smash{\begin{tabular}[t]{c}Ground Truth\end{tabular}}}}%
    \put(0,0){\includegraphics[width=\unitlength,page=1]{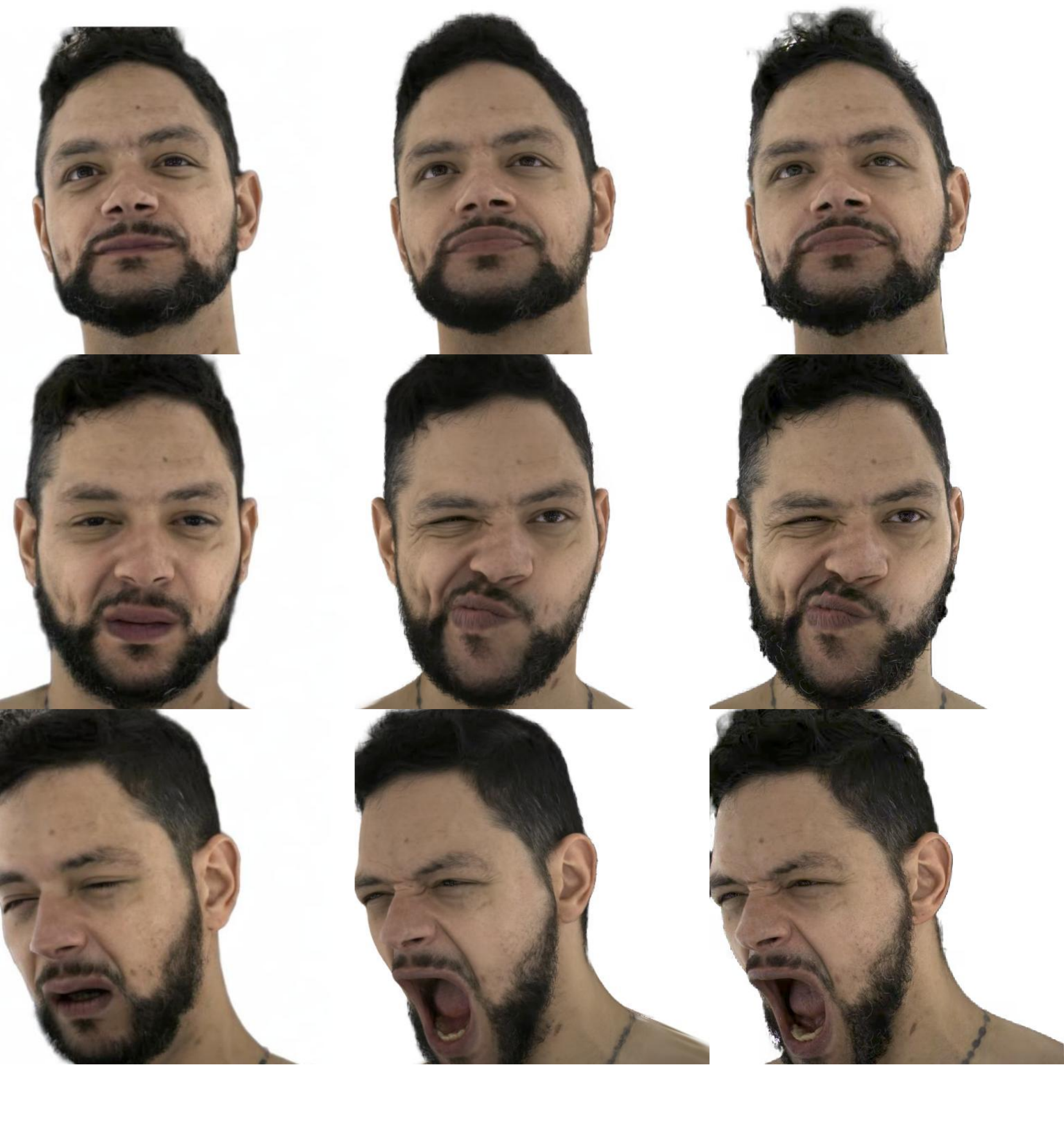}}%
  \end{picture}%
\endgroup%

    \caption{Novel view synthesis comparison against CAP4D on Ava-256.}
    \label{fig:nvs_cap4d}
\end{figure}

\begin{figure}[t]
    \centering
    \def\svgwidth{\linewidth}
    \begingroup%
  \makeatletter%
  \providecommand\color[2][]{%
    \errmessage{(Inkscape) Color is used for the text in Inkscape, but the package 'color.sty' is not loaded}%
    \renewcommand\color[2][]{}%
  }%
  \providecommand\transparent[1]{%
    \errmessage{(Inkscape) Transparency is used (non-zero) for the text in Inkscape, but the package 'transparent.sty' is not loaded}%
    \renewcommand\transparent[1]{}%
  }%
  \providecommand\rotatebox[2]{#2}%
  \newcommand*\fsize{\dimexpr\f@size pt\relax}%
  \newcommand*\lineheight[1]{\fontsize{\fsize}{#1\fsize}\selectfont}%
  \ifx\svgwidth\undefined%
    \setlength{\unitlength}{1152.00006921bp}%
    \ifx\svgscale\undefined%
      \relax%
    \else%
      \setlength{\unitlength}{\unitlength * \real{\svgscale}}%
    \fi%
  \else%
    \setlength{\unitlength}{\svgwidth}%
  \fi%
  \global\let\svgwidth\undefined%
  \global\let\svgscale\undefined%
  \makeatother%
  \small
  \begin{picture}(1,1.30531649)%
    \lineheight{1}%
    \setlength\tabcolsep{0pt}%
    \put(0,0){\includegraphics[width=\unitlength,page=1]{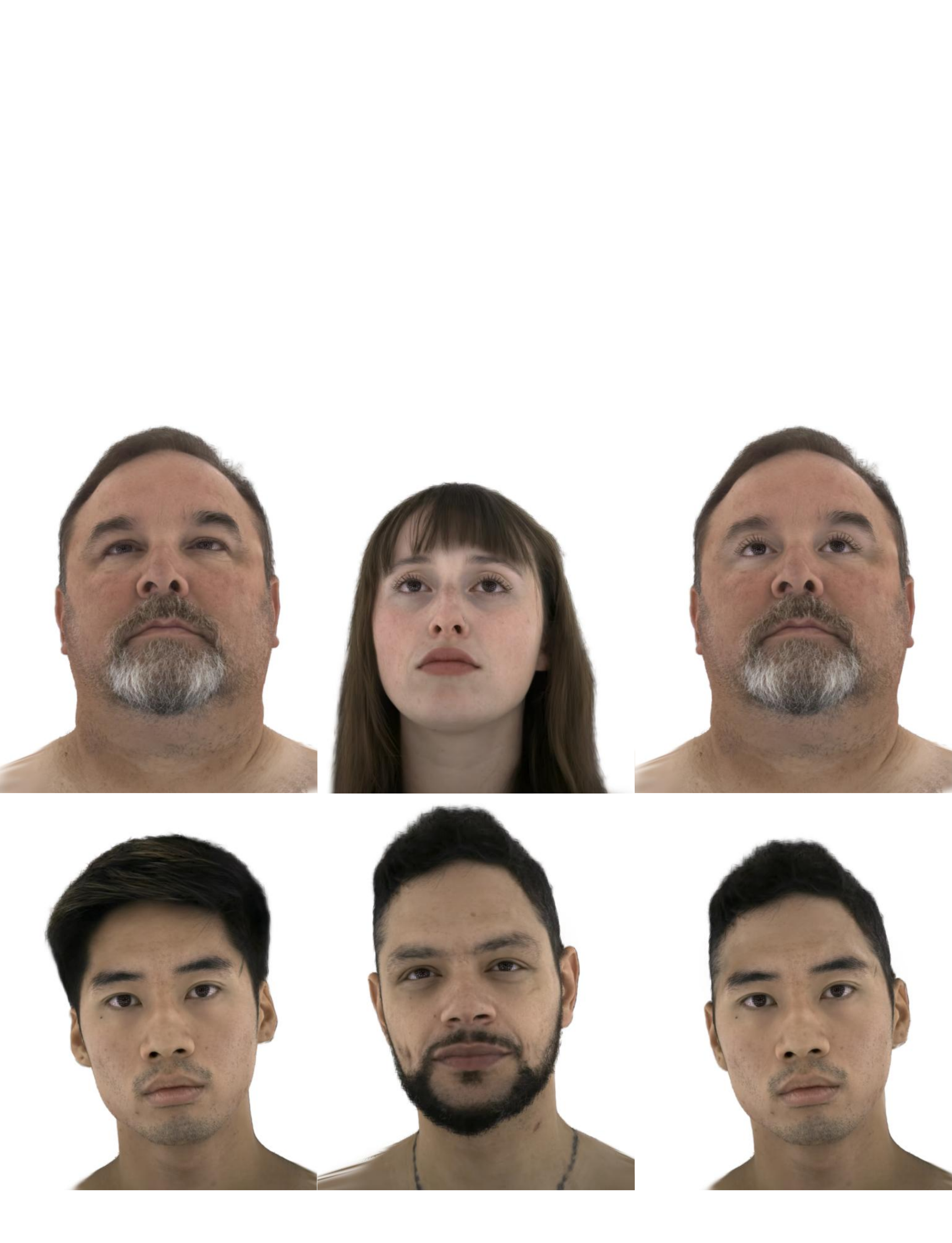}}%
    \put(0.16666669,0.0091167){\color[rgb]{0,0,0}\makebox(0,0)[t]{\lineheight{0}\smash{\begin{tabular}[t]{c}Source Identity\end{tabular}}}}%
    \put(0.49942319,0.0091167){\color[rgb]{0,0,0}\makebox(0,0)[t]{\lineheight{0}\smash{\begin{tabular}[t]{c}Target Attribute\end{tabular}}}}%
    \put(0.83333331,0.0091167){\color[rgb]{0,0,0}\makebox(0,0)[t]{\lineheight{0}\smash{\begin{tabular}[t]{c}Result\end{tabular}}}}%
    \put(0,0){\includegraphics[width=\unitlength,page=2]{editing_b.pdf}}%
  \end{picture}%
\endgroup%

    \caption{Additional semantic editing results. From top to bottom: Transferring beard and lips, eyes, and hairstyle.}
    \label{fig:editing_b}
\end{figure}

\begin{figure*}[t]
    \centering
    \def\svgwidth{0.87\linewidth}
    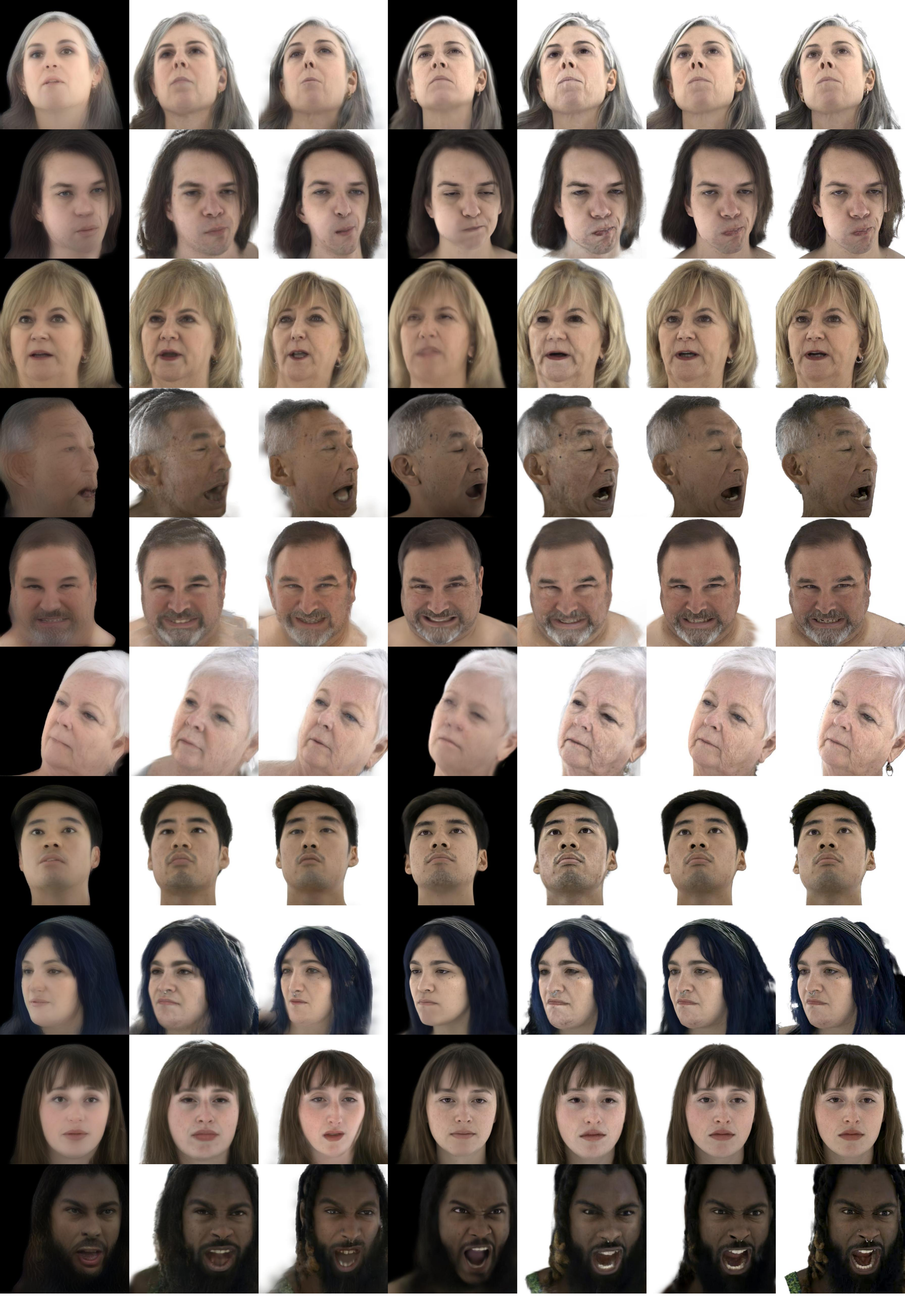
    \caption{Additional novel view synthesis results on Ava-256~\cite{martinez2024codec}.}
    \label{fig:nvs_supmat}
\end{figure*}

\begin{figure*}[t]
    \centering
    \def\svgwidth{\linewidth}
    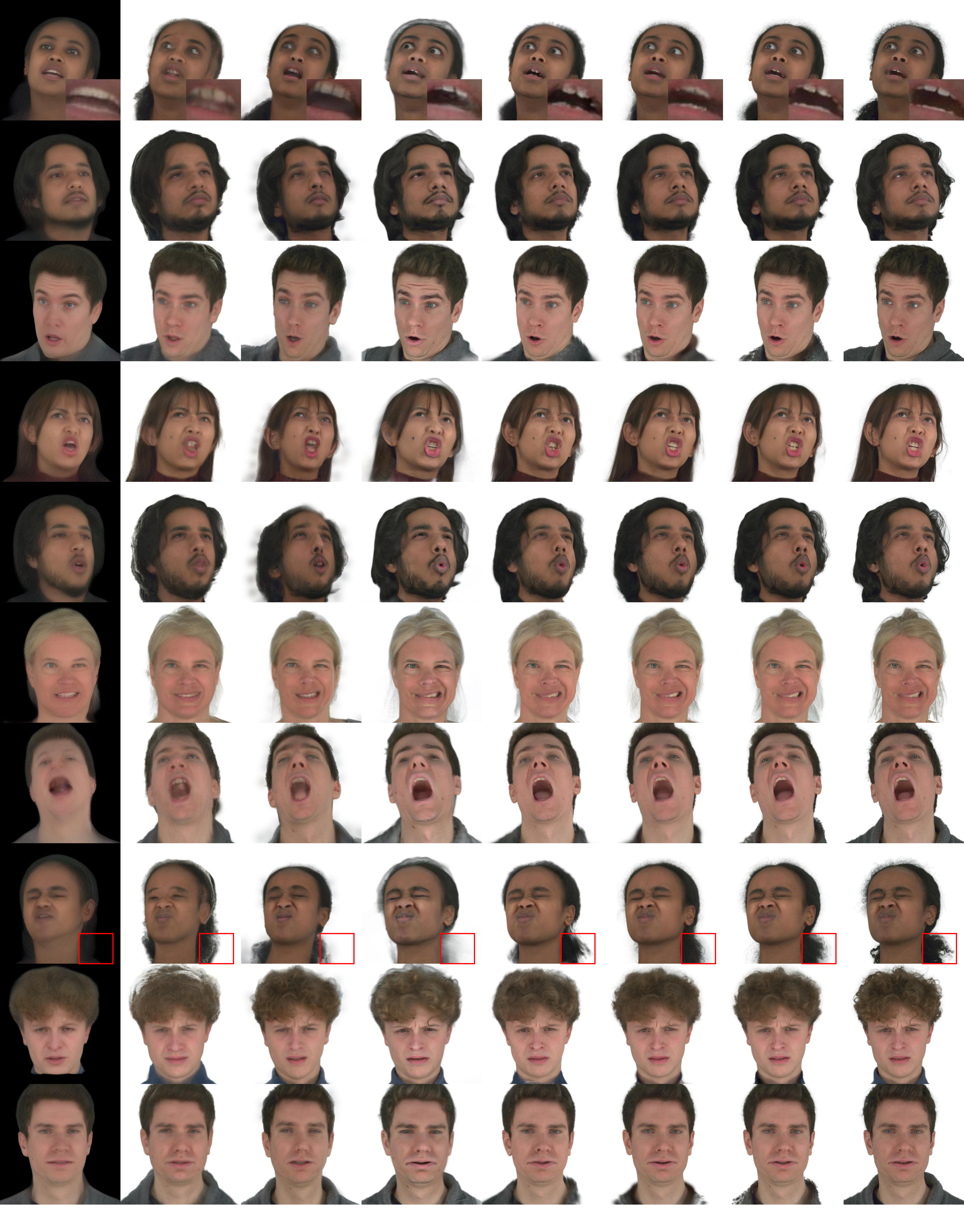
    \caption{Additional novel view synthesis results on NeRSemble~\cite{kirschstein2023nersemble}. Ours (Ava) / Ours (NeRSemble) are trained on Ava-256~\cite{martinez2024codec} and NeRSemble only, respectively.}
    \label{fig:nvs_nersemble_big}
\end{figure*}

\Cref{fig:nvs_cap4d} presents the qualitative comparison with CAP4D on the one intersecting validation subject. 
We observe that our method predicts reconstructions with better identity preservation and expression fidelity. 
\Cref{fig:nvs_supmat} and \Cref{fig:nvs_nersemble_big} provide additional comparisons with the remaining baselines on samples from Ava-256 and NeRSemble respectively. 
As discussed in the main paper, our method exhibits superior synthesis quality.

\subsection{Additional Ablations}
\label{sec:supp_ablations}

\begin{table}  %
\resizebox{\linewidth}{!}{%
\begin{tabular}{ll|rrrrrr}
\toprule 
 $V_\text{TEMPEH}$& $V_\text{\modelname}$&LPIPS $\downarrow$ & CSIM $\uparrow$ & PSNR $\uparrow$ & SSIM $\uparrow$ & L1 $\downarrow$ & L2 $\downarrow$ \\
\midrule
12 & 2 & 0.213 & 0.866 & 20.503 & 0.795 & 0.042 & 0.013 \\
12 & 4 & 0.212 & 0.865 & 21.078 & 0.798 & 0.039 & 0.012 \\
12 & 8 & 0.195 & 0.907 & 22.350 & 0.816 & 0.034 & 0.010 \\
12 & 12 & \cellcolor{tabsecond}0.187 & \cellcolor{tabsecond}0.918 & \cellcolor{tabsecond}23.032 & \cellcolor{tabsecond}0.825 & \cellcolor{tabfirst}0.032 & \cellcolor{tabfirst}0.009 \\
12 & 16 & \cellcolor{tabfirst}0.182 & \cellcolor{tabfirst}0.923 & \cellcolor{tabfirst}23.367 & \cellcolor{tabfirst}0.831 & \cellcolor{tabfirst}0.032 & \cellcolor{tabfirst}0.009 \\
\midrule
2 & 2 & 0.261 & 0.689 & 15.996 & 0.731 & 0.071 & 0.033 \\
4 & 4 & 0.230 & 0.798 & 19.289 & 0.775 & 0.047 & 0.017 \\
8 & 8 & 0.198 & 0.903 & 22.076 & 0.812 & 0.035 & 0.010 \\
12 & 12 & \cellcolor{tabsecond}0.187 & \cellcolor{tabsecond}0.918 & \cellcolor{tabsecond}23.032 & \cellcolor{tabsecond}0.825 & \cellcolor{tabfirst}0.032 & \cellcolor{tabfirst}0.009 \\
16 & 16 & \cellcolor{tabfirst}0.182 & \cellcolor{tabfirst}0.924 & \cellcolor{tabfirst}23.265 & \cellcolor{tabfirst}0.831 & \cellcolor{tabfirst}0.032 & \cellcolor{tabfirst}0.009 \\
\bottomrule
\end{tabular}
}
\caption{Quantitative ablation of the number of input views to \modelname evaluated on Ava-256. We evaluate two scenarios: $i)$ Changing the number of input views to \modelname while keeping the number of inputs to the coarse mesh registration model (TEMPEH) at the default ($V=12$). $ii)$ Changing the number of input views for both TEMPEH and \modelname.}
\label{tab:nvs_ablation_nviews}
\end{table}

\begin{figure*}[t]
    \centering
    \def\svgwidth{\linewidth}
    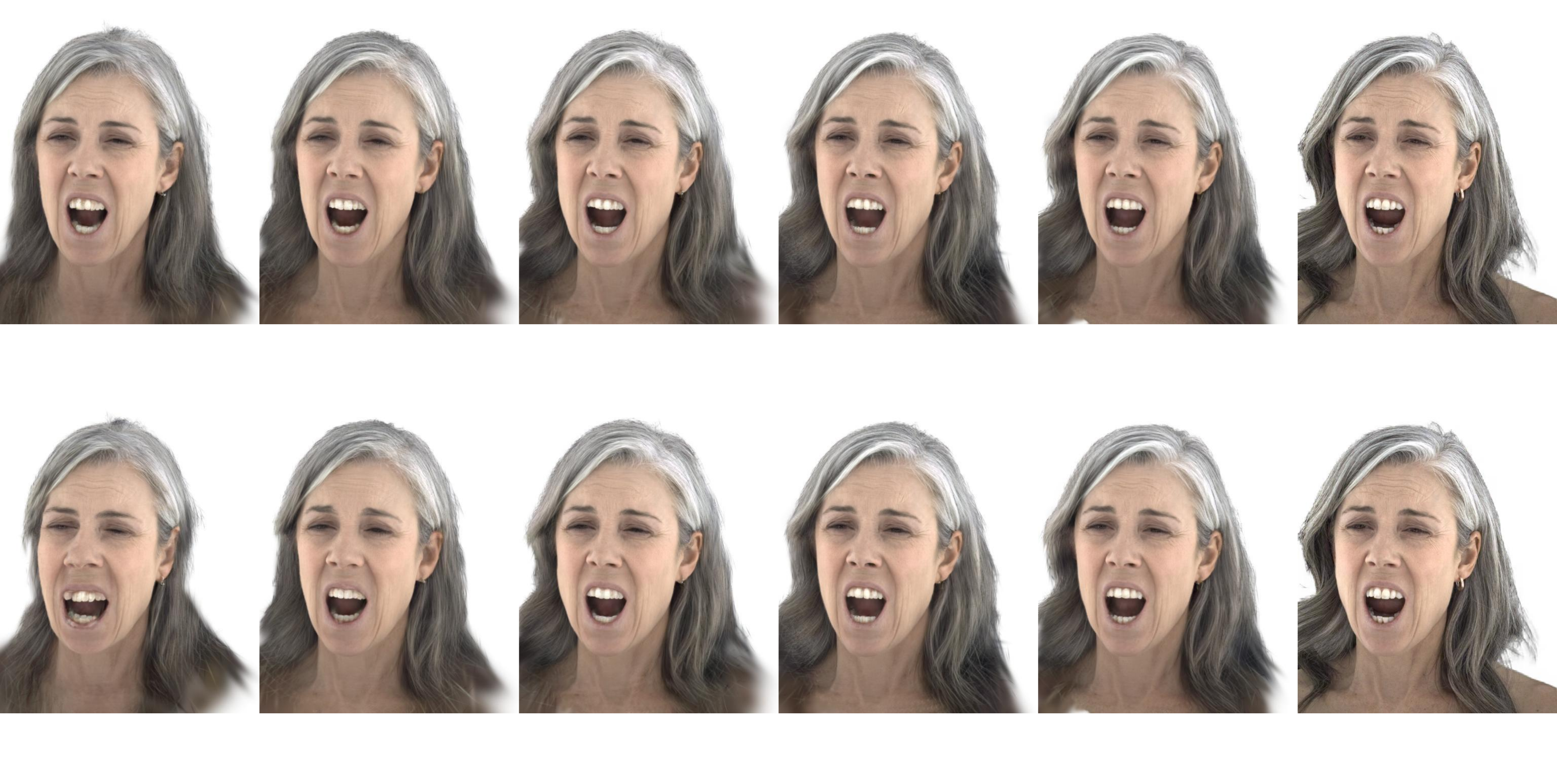
    \caption{Qualitative ablation study for the number of input views to \modelname. We evaluate two scenarios. Top: Changing the number of input views to \modelname while keeping the number of inputs to the coarse mesh registration model (TEMPEH) at the default ($V=12$). Bottom: Changing the number of input views for both TEMPEH and \modelname.}
    \label{fig:nvs_ablation_nviews_qual}
\end{figure*}

\begin{figure*}[!t]
    \centering
    \def\svgwidth{\linewidth}
    \begingroup%
  \makeatletter%
  \providecommand\color[2][]{%
    \errmessage{(Inkscape) Color is used for the text in Inkscape, but the package 'color.sty' is not loaded}%
    \renewcommand\color[2][]{}%
  }%
  \providecommand\transparent[1]{%
    \errmessage{(Inkscape) Transparency is used (non-zero) for the text in Inkscape, but the package 'transparent.sty' is not loaded}%
    \renewcommand\transparent[1]{}%
  }%
  \providecommand\rotatebox[2]{#2}%
  \newcommand*\fsize{\dimexpr\f@size pt\relax}%
  \newcommand*\lineheight[1]{\fontsize{\fsize}{#1\fsize}\selectfont}%
  \ifx\svgwidth\undefined%
    \setlength{\unitlength}{1919.99994233bp}%
    \ifx\svgscale\undefined%
      \relax%
    \else%
      \setlength{\unitlength}{\unitlength * \real{\svgscale}}%
    \fi%
  \else%
    \setlength{\unitlength}{\svgwidth}%
  \fi%
  \global\let\svgwidth\undefined%
  \global\let\svgscale\undefined%
  \makeatother%
  \small
  \begin{picture}(1,0.23865746)%
    \lineheight{1}%
    \setlength\tabcolsep{0pt}%
    \put(0.69999999,0.00115747){\color[rgb]{0,0,0}\makebox(0,0)[t]{\lineheight{0}\smash{\begin{tabular}[t]{c}$k_{\mathcal{T},\text{img}}=150$\end{tabular}}}}%
    \put(0.09999994,0.00115747){\color[rgb]{0,0,0}\makebox(0,0)[t]{\lineheight{0}\smash{\begin{tabular}[t]{c}$k_{\mathcal{T},\text{img}}=25$\end{tabular}}}}%
    \put(0.50000004,0.00115745){\color[rgb]{0,0,0}\makebox(0,0)[t]{\lineheight{0}\smash{\begin{tabular}[t]{c}Ours\end{tabular}}}}%
    \put(0.30000003,0.00115747){\color[rgb]{0,0,0}\makebox(0,0)[t]{\lineheight{0}\smash{\begin{tabular}[t]{c}$k_{\mathcal{T},\text{img}}=50$\end{tabular}}}}%
    \put(0.90000001,0.00115747){\color[rgb]{0,0,0}\makebox(0,0)[t]{\lineheight{0}\smash{\begin{tabular}[t]{c}Ground Truth\end{tabular}}}}%
    \put(0,0){\includegraphics[width=\unitlength,page=1]{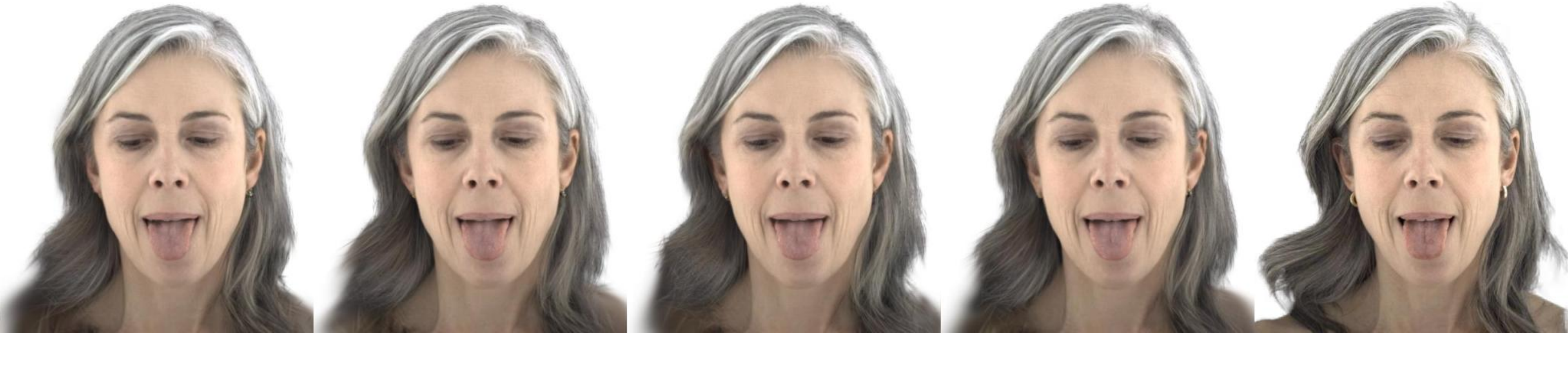}}%
  \end{picture}%
\endgroup%

    \caption{Qualitative ablation study for $k_{\mathcal{T}, \text{img}}$, i.e., the number of image tokens that each UV token attends to in the registration-guided attention blocks. The default value is $k_{\mathcal{T}, \text{img}}=100$.}
    \label{fig:nvs_ablation_k_img_qual}
\end{figure*}

\begin{figure}[t]
    \centering
    \begin{center}
        \resizebox{\linewidth}{!}{\input{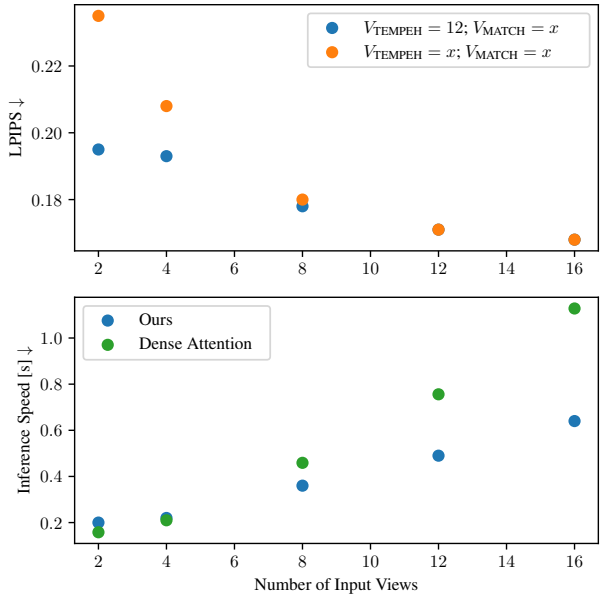}}
    \end{center}
    \caption{Top: Quantitative ablation study for the number of input views to \modelname on Ava-256. We evaluate two scenarios: $i)$ Changing the number of input views to \modelname while keeping the number of inputs to the coarse mesh registration model (TEMPEH) at the default ($V=12$). $ii)$ Changing the number of input views for both TEMPEH and \modelname. Bottom: Inference speed comparison between our model with the novel registration-guided attention versus a version with dense attention across all UV and image tokens. }
    \label{fig:nvs_ablation_nviews_quant}
\end{figure}
\paragraph{Number of input images.}
\Cref{fig:nvs_ablation_nviews_qual} qualitatively evaluates the impact of the number of input views on the synthesis result. 
We conduct two lines of experiments: $i)$ Keeping the number of input images to the coarse mesh registration model (TEMPEH) at the default ($V=12$) while only changing the number of input views to \modelname. This evaluates the actual impact of the number of input views on our method in isolation. 
$ii)$ TEMPEH and \modelname receive the same number of input images. This is the more realistic scenario, but entangles the sensitivity of TEMPEH to few input images with \modelname's.

We find that \modelname is highly robust to few input images and can generate plausible reconstructions even for two input images, assuming high-quality geometry initialization. 
However, TEMPEH's geometry prediction degrades significantly for two views, resulting in a low-quality reconstruction for the combined scenario. 
If TEMPEH and \modelname receive the same number of input views, starting from four images, plausible results are produced.
Fine details improve as more input views are added. 
This is confirmed by \Cref{tab:nvs_ablation_nviews} and \Cref{fig:nvs_ablation_nviews_quant} (top), which show improving LPIPS scores as the number of views increases. 
\Cref{fig:nvs_ablation_nviews_qual} (bottom) reports the inference speed. As discussed in the main paper, while the computational complexity scales quadratically with the number of input images for dense attention between all UV and image tokens, our method's complexity increases only linearly. 
Especially for high numbers of input images, this results in a considerable improvement of inference speed ($1.8\times$ acceleration compared to dense attention for 16 input images). 
We found 12 input images to be a good compromise between inference speed and synthesis quality, running at a framerate of 2 fps. 

\paragraph{Registration-guided attention context length.}
We ablate the effect of $k_{\mathcal{T}, \text{img}}$, i.e., the number of image tokens that each UV token attends to in the registration-guided attention blocks. 
The quantitative comparison in \Cref{tab:nvs_ablation_k_img} shows minor improvements as we decrease $k_{\mathcal{T}, \text{img}}$. However, we did not observe pronounced qualitative differences as shown in \Cref{fig:nvs_ablation_k_img_qual}.

\begin{figure}[t]
    \centering
    \def\svgwidth{\linewidth}
    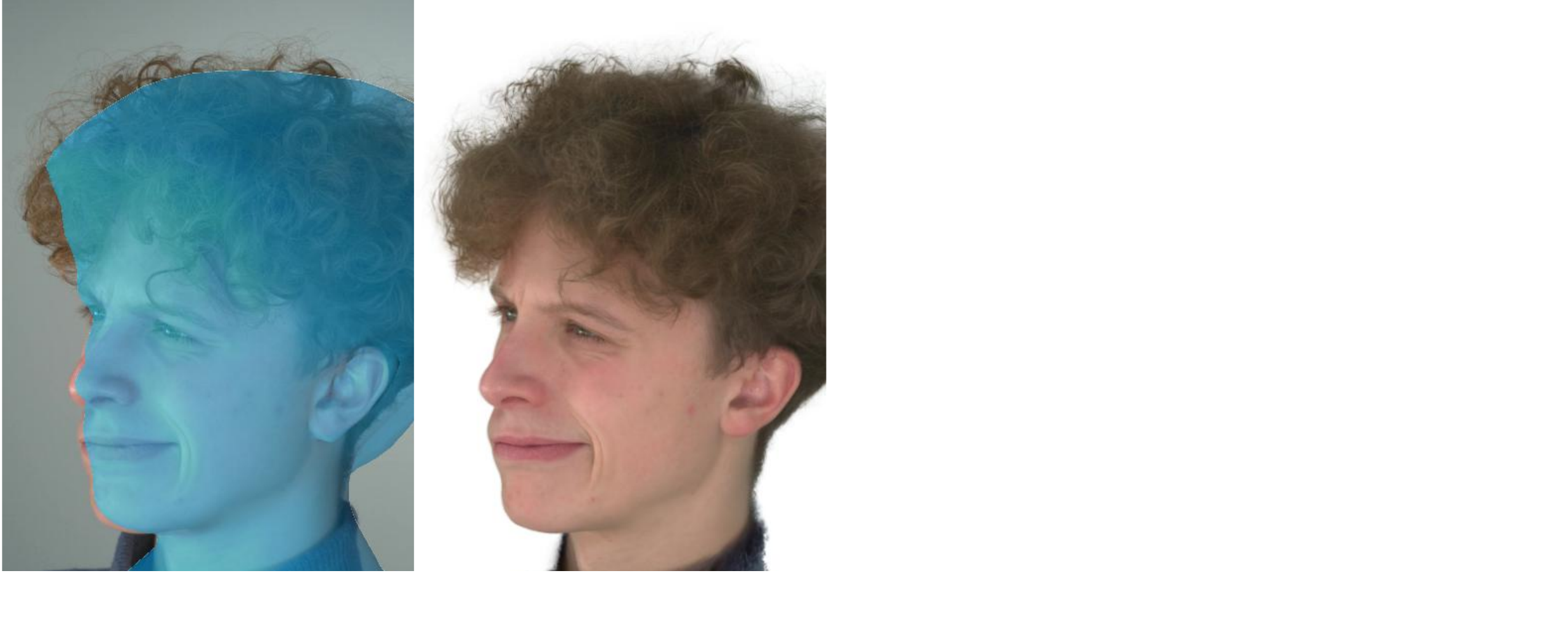
    \caption{Robustness to errors in the coarse TEMPEH mesh.
    }
    \label{fig:robustness_to_coarse_mesh}
\end{figure}
\paragraph{Robustness to coarse mesh errors.}
\modelname uses a coarse mesh estimated by TEMPEH as initialization. 
In practice, TEMPEH exhibits moderate inaccuracies, which \modelname can recover from, see \Cref{fig:robustness_to_coarse_mesh}~(left), producing plausible results.
When perturbing the coarse mesh by constant vertex offsets $\Delta x$, \Cref{fig:robustness_to_coarse_mesh}~(right), artifacts only appear for $\Delta x \geq 20\text{mm}$, which is $7\times$ higher than the average point-to-surface distance of TEMPEH. 

\paragraph{Image token self attention.}
\modelname performs self-attention between the image tokens of each image to enable image-space signal processing inside the grouped attention blocks. We found that this self-attention can be skipped without harming the model performance while increasing the inference speed by 8\%.

\begin{table}  %
\resizebox{\linewidth}{!}{%
\begin{tabular}{l|rrrrrr}
\toprule 
 $k_{\mathcal{T}, \text{img}}$&LPIPS $\downarrow$&CSIM $\uparrow$&PSNR $\uparrow$&SSIM $\uparrow$&L1 $\downarrow$&L2 $\downarrow$\\
\midrule
25 & \cellcolor{tabsecond}0.184 & \cellcolor{tabfirst}0.926 & 22.908 & \cellcolor{tabfirst}0.830 & \cellcolor{tabsecond}0.032 & \cellcolor{tabsecond}0.009 \\
50 & \cellcolor{tabfirst}0.183 & \cellcolor{tabsecond}0.924 & \cellcolor{tabfirst}23.164 & \cellcolor{tabsecond}0.830 & \cellcolor{tabfirst}0.031 & \cellcolor{tabfirst}0.008 \\
100 & 0.187 & 0.918 & \cellcolor{tabsecond}23.032 & 0.825 & \cellcolor{tabsecond} 0.032 & \cellcolor{tabsecond}0.009 \\
150 & 0.185 & 0.919 & 22.951 & 0.826 & \cellcolor{tabsecond} 0.032 & \cellcolor{tabsecond}0.009 \\
\bottomrule
\end{tabular}
}
\caption{Quantitative ablation study for $k_{\mathcal{T}, \text{img}}$, i.e., the number of image tokens that each UV token attends to in the registration-guided attention blocks. The evaluations were performed on the Ava-256 dataset. The default value is $k_{\mathcal{T}, \text{img}}=100$.}
\label{tab:nvs_ablation_k_img}
\end{table}

\section{In-the-wild application.}
\label{sec:itw}
\begin{figure}[t]
    \centering
    \def\svgwidth{\linewidth}
    \begingroup%
  \makeatletter%
  \footnotesize
  \providecommand\color[2][]{%
    \errmessage{(Inkscape) Color is used for the text in Inkscape, but the package 'color.sty' is not loaded}%
    \renewcommand\color[2][]{}%
  }%
  \providecommand\transparent[1]{%
    \errmessage{(Inkscape) Transparency is used (non-zero) for the text in Inkscape, but the package 'transparent.sty' is not loaded}%
    \renewcommand\transparent[1]{}%
  }%
  \providecommand\rotatebox[2]{#2}%
  \newcommand*\fsize{\dimexpr\f@size pt\relax}%
  \newcommand*\lineheight[1]{\fontsize{\fsize}{#1\fsize}\selectfont}%
  \ifx\svgwidth\undefined%
    \setlength{\unitlength}{1980.98340613bp}%
    \ifx\svgscale\undefined%
      \relax%
    \else%
      \setlength{\unitlength}{\unitlength * \real{\svgscale}}%
    \fi%
  \else%
    \setlength{\unitlength}{\svgwidth}%
  \fi%
  \global\let\svgwidth\undefined%
  \global\let\svgscale\undefined%
  \makeatother%
  \begin{picture}(1,0.74553044)%
    \lineheight{1}%
    \setlength\tabcolsep{0pt}%
    \put(0,0){\includegraphics[width=\unitlength,page=1]{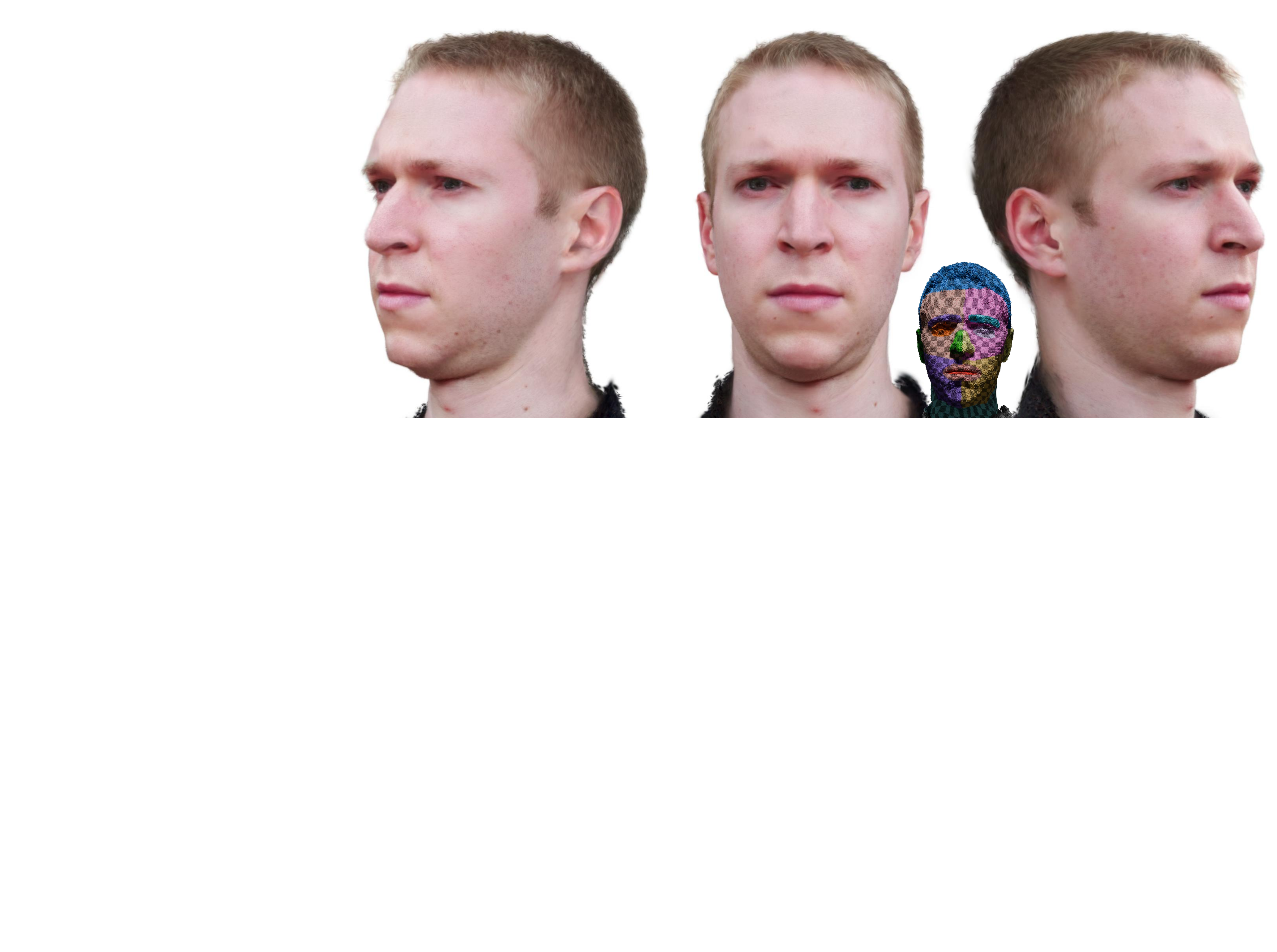}}%
    \put(0.13697561,0.38325961){\color[rgb]{0,0,0}\makebox(0,0)[t]{\lineheight{0}\smash{\begin{tabular}[t]{c}4 of 12 Input Images\end{tabular}}}}%
    \put(0.63650313,0.38325961){\color[rgb]{0,0,0}\makebox(0,0)[t]{\lineheight{0}\smash{\begin{tabular}[t]{c}\modelname Prediction\end{tabular}}}}%
    \put(0,0){\includegraphics[width=\unitlength,page=2]{itw_demo_single_image_combined_v1.pdf}}%
    \put(0.1369756,0.00530161){\color[rgb]{0,0,0}\makebox(0,0)[t]{\lineheight{0}\smash{\begin{tabular}[t]{c}Single Input Image\end{tabular}}}}%
    \put(0.63650314,0.00530161){\color[rgb]{0,0,0}\makebox(0,0)[t]{\lineheight{0}\smash{\begin{tabular}[t]{c}\modelname Prediction\end{tabular}}}}%
    \put(0,0){\includegraphics[width=\unitlength,page=3]{itw_demo_single_image_combined_v1.pdf}}%
  \end{picture}%
\endgroup%

    \caption{In-the-wild (top) and single-image (bottom) results.}
    \label{fig:itw_demo__single_image}
\end{figure}
While \modelname was trained on calibrated studio-captures with uniform lighting and known camera parameters, we found that it generalizes to in-the-wild captures and yields high-quality reconstructions, see \Cref{fig:itw_demo__single_image}~(top). 
The input images were captured with an off-the-shelf camera in an outdoor environment, and we used COLMAP~\cite{fisher2021colmap} to estimate the camera parameters.

\section{Single-image inference.}
\label{sec:single-image}
\modelname is not trained to hallucinate unobserved regions and shows artifacts for two input images only (see \Cref{fig:nvs_ablation_nviews_qual}). 
However, we can follow FaceLift~\cite{lyu2025facelift} to generate additional views from a single input image with the 2D prior of CAP4D and input these to \modelname.
\Cref{fig:itw_demo__single_image} (bottom) shows that this yields high-quality reconstructions. 

\section{Additional Results for Interpolation, Editing, and Expression Transfer}
\label{sec:supp_editing}
\begin{figure}[t]
    \centering
    \def\svgwidth{\linewidth}
    \begingroup%
  \makeatletter%
  \providecommand\color[2][]{%
    \errmessage{(Inkscape) Color is used for the text in Inkscape, but the package 'color.sty' is not loaded}%
    \renewcommand\color[2][]{}%
  }%
  \providecommand\transparent[1]{%
    \errmessage{(Inkscape) Transparency is used (non-zero) for the text in Inkscape, but the package 'transparent.sty' is not loaded}%
    \renewcommand\transparent[1]{}%
  }%
  \providecommand\rotatebox[2]{#2}%
  \newcommand*\fsize{\dimexpr\f@size pt\relax}%
  \newcommand*\lineheight[1]{\fontsize{\fsize}{#1\fsize}\selectfont}%
  \ifx\svgwidth\undefined%
    \setlength{\unitlength}{1151.9999827bp}%
    \ifx\svgscale\undefined%
      \relax%
    \else%
      \setlength{\unitlength}{\unitlength * \real{\svgscale}}%
    \fi%
  \else%
    \setlength{\unitlength}{\svgwidth}%
  \fi%
  \global\let\svgwidth\undefined%
  \global\let\svgscale\undefined%
  \makeatother%
  \small
  \begin{picture}(1,1.1900145)%
    \lineheight{1}%
    \setlength\tabcolsep{0pt}%
    \put(0.50000001,0.00911665){\color[rgb]{0,0,0}\makebox(0,0)[t]{\lineheight{0}\smash{\begin{tabular}[t]{c}Target Expression\end{tabular}}}}%
    \put(0.16666667,0.00911665){\color[rgb]{0,0,0}\makebox(0,0)[t]{\lineheight{0}\smash{\begin{tabular}[t]{c}Source Identity\end{tabular}}}}%
    \put(0.83333333,0.00911665){\color[rgb]{0,0,0}\makebox(0,0)[t]{\lineheight{0}\smash{\begin{tabular}[t]{c}Result\end{tabular}}}}%
    \put(0,0){\includegraphics[width=\unitlength,page=1]{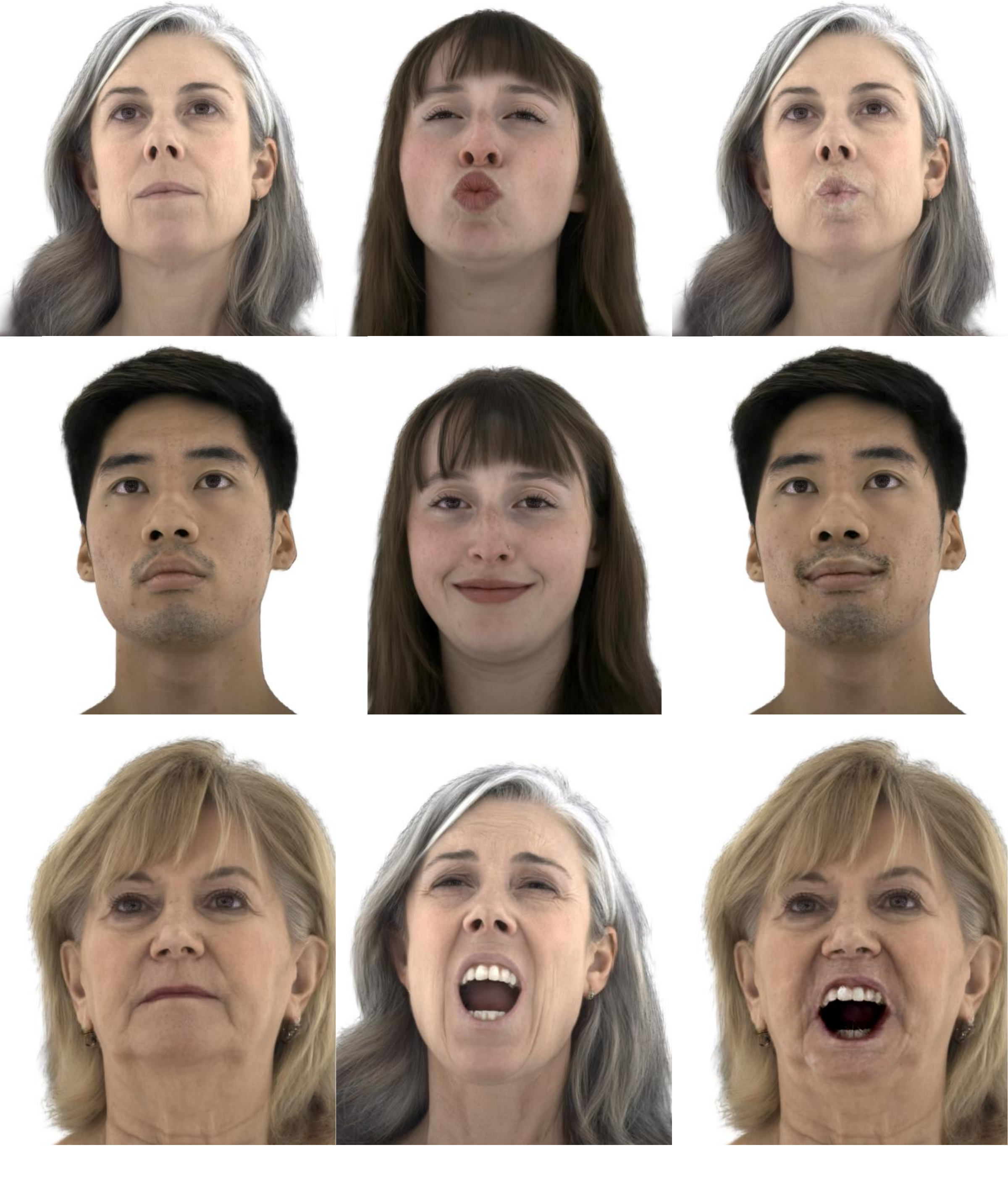}}%
  \end{picture}%
\endgroup%

    \caption{Additional expression transfer results. Note that we only aim to transfer the oral expression and do not apply any modifications to other regions, e.g., the eyes.}
    \label{fig:expr_transfer_b}
\end{figure}

\begin{figure*}[t]
    \centering
    \def\svgwidth{\linewidth}
    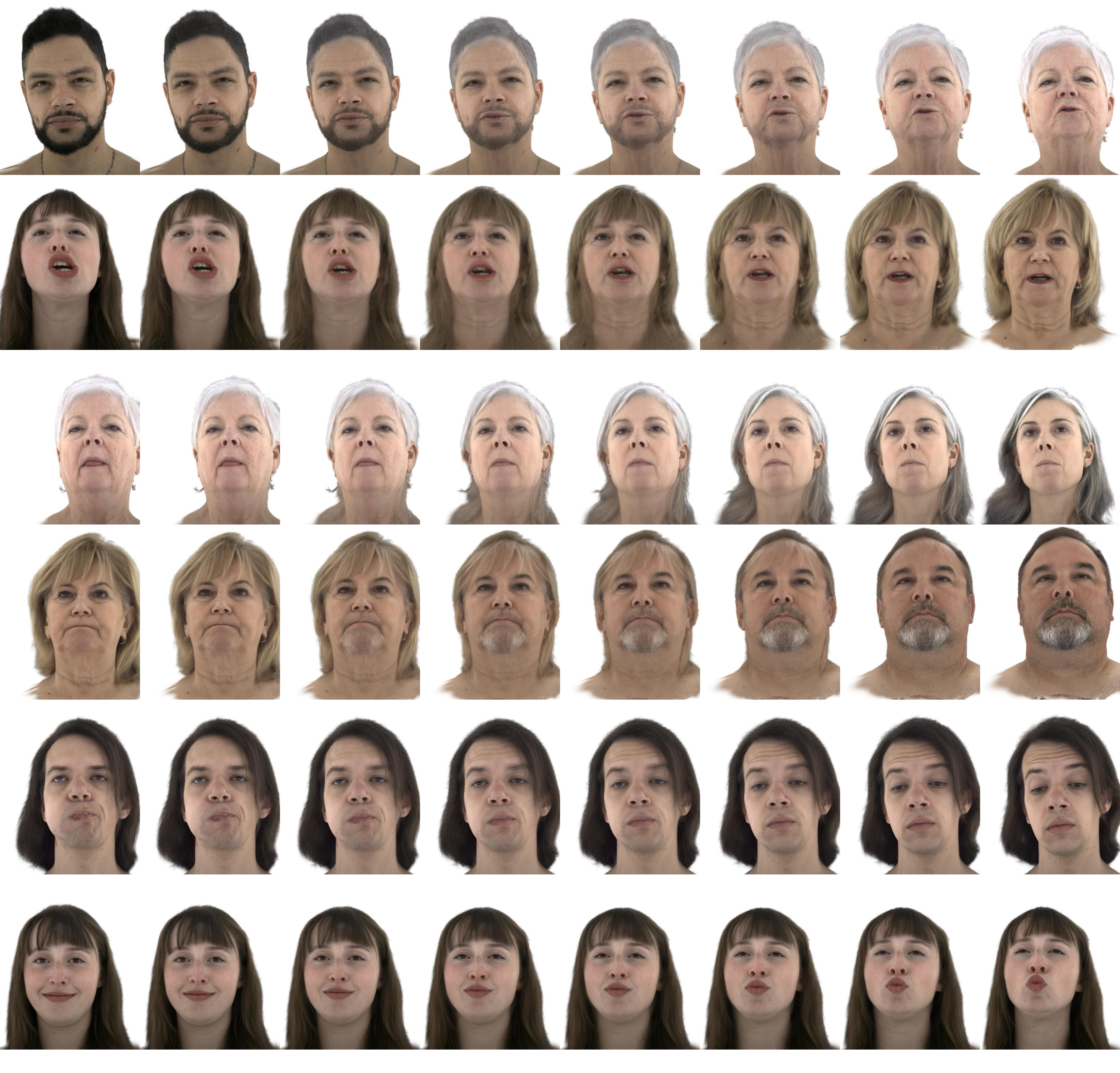
    \caption{Additional interpolation results. $\gamma$ denotes the interpolation factor.}
    \label{fig:interpolation_full}
\end{figure*}

\Cref{fig:interpolation_full}, \Cref{fig:editing_b}, \Cref{fig:expr_transfer_b} present further results for interpolation, semantic editing, and expression transfer, respectively. 
We observe smooth interpolations between samples and plausible editing results for swapping beard, eyes, and hairstyle.
As discussed in the main paper, the arithmetic expression transfer approach, where the residual of Gaussian maps for an expressive and a neutral frame of a target subject is added to the neutral reconstruction of a source identity, can result in uncanny results for extreme expressions and dissimilar identities. 
A less simplistic method, e.g., a conditional VAE~\cite{martinez2024codec}, would be a more suitable choice for this challenging task. 

\section{Quantitative Correspondence Evaluation.}
\label{sec:quant_correspondence_eval}

\begin{table}  %
\resizebox{\linewidth}{!}{%
\begin{tabular}{l|cccccc}
\toprule 
Corresp. Dist. [mm] $\downarrow$  & Full Head & Face & Ears & Eyes & Mouth & Scalp \\
\midrule
TEMPEH~\cite{bolkart2023tempeh} / Ours & 8.9 / \textbf{8.0} & 5.4 / \textbf{4.8}& 10.8 / \textbf{9.1} & 2.5 / \textbf{2.1}& 2.8 /  \textbf{2.7}& 13.5 / \textbf{12.5}\\
\bottomrule
\end{tabular}
}
\caption{Quantitative correspondence evaluation.\vspace{-0.1cm}}
\label{tab:quant_correspondence}
\end{table}

We quantify the semantic correspondence of \modelname’s predictions using Ava-256's ground truth mesh registrations. 
\Cref{tab:quant_correspondence} reports the Euclidean distance between the center of each predicted Gaussian and its corresponding target location obtained through barycentric interpolation on 1,000 samples. 
The same interpolation is done to evaluate TEMPEH's results.
We find that \modelname produces superior correspondence. 

\section{Additional Material for the Subject-Specific Avatars}
\label{sec:supp_avatars}
\subsection{Detailed Avatar Creation Procedure}
\begin{figure}[t]
    \centering
    \includegraphics[width=\linewidth]{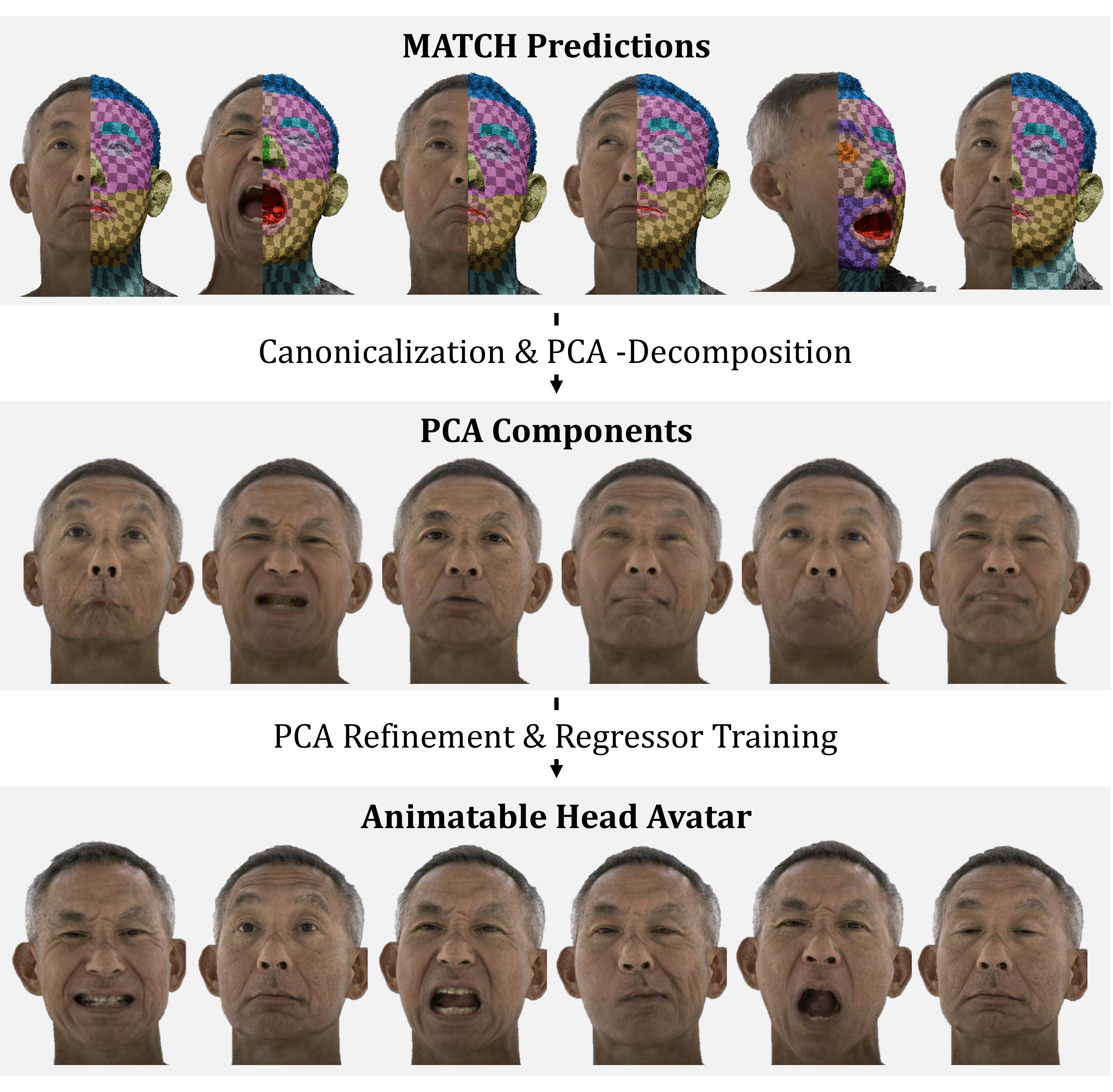}
    \caption{Procedure to create subject-specific head avatars from a sequence of Gaussian splat textures predicted by \modelname.}
    \label{fig:avatar_creation}
\end{figure}

This section illustrates the changes applied to GEM's~\cite{zielonka2025gem} procedure to create a lightweight animatable avatar from a set of Gaussian splat textures predicted by \modelname. 
Ablations on the effect of the individual changes are presented in \Cref{fig:dist_ablation} and \Cref{tab:distillation_ablation}, which are discussed in \Cref{sec:supp_dist_ablations}.
\Cref{fig:avatar_creation} provides an overview of the resulting procedure. 

\emph{i) Skip Tracking \& CNN-based Avatar Training:} Since \modelname directly predicts Gaussian splats that are in correspondence across frames, we can skip the time-expensive procedure of tracking and CNN-based head avatar training, which drastically reduces the time to create a lightweight head avatar (see \Cref{tab:supp_timing}). 
Since the reconstruction of the PCA basis requires unposed Gaussians in a canonical space, we have to unpose \modelname's predictions. 
To this end, we extract the Ava-256 mesh by sampling the texture of predicted 3D Gaussian locations at the template vertices' UV coordinates. 
We then convert the Ava-256 mesh to the topology of FLAME~\cite{li2017flame}, a publicly available 3D morphable model (3DMM), using a fixed mapping of vertex locations. 
The 3DMM parameters are obtained by optimizing FLAME's vertices against our vertex predictions using a Huber loss~\cite{huber1992robust}.
Finally, we can use the obtained FLAME pose parameters to apply inverse linear blend skinning to transform the Gaussians predicted by \modelname into an unposed canonical space on which we perform the PCA decomposition. 
Note that during this unposing operation, the jaw articulation is neutralized as well. For this reason, we train GEM's expression encoder to also predict the jaw pose in addition to the Eigen-coefficients. 
To reduce the compute cost and memory requirements of the PCA decomposition, we use a version of \modelname that predicts Gaussian textures with a reduced resolution of $512\times512$. 

\emph{ii) Modality-agnostic PCA:} GEM creates separate PCAs for each of the Gaussian's modalities (rotation, position, opacity, and scale). 
However, we found that this formulation misses crucial correlations between the modalities (e.g., raised eyebrows should correlate with color changes in wrinkles on the forehead). 
This is resolved by modelling all Gaussian modalities in a joint PCA. 

\emph{iii) Enable dynamic colors:} GEM disables dynamic color changes to promote semantic correspondence of Gaussians across frames.  
For \modelname, however, this is neither feasible nor practical, since it must predict dynamically changing colors to reconstruct the appearance of different subjects, and intrinsically exhibits high semantic correspondence across subjects and frames. 
As such, we drop the constraint of static colors during the PCA reconstruction and refinement. 
The only exception to this is the interior of the mouth cavity. 
Since the ground truth mesh registrations on the Ava-256 dataset simplify the oral cavity as planar surfaces between the lips, the semantic correspondence of \modelname's predictions in this area is limited. Plausible reconstructions are achieved through intricately changing colors, opacities, and scales. 
We found that na\"ively using the lightweight expression MLP to predict these complex dynamics yields test-time artifacts. 
To alleviate this problem, we fix the colors, scales, and opacities of the Gaussians in the oral cavity to their mean calculated across all training frames. 

\emph{iv) Mean Refinement:} GEM only refines the PCA basis vectors $\mathbf{B}_i$ against the target images using photometric losses. We found it beneficial to also refine the PCA means $\boldsymbol{\mu}_i$ during that stage. 

\subsection{Detailed Baseline Description}
\label{sec:supp_avatar_baselines}
We compare our avatars with the optimization-based methods GaussianAvatars~\cite{qian2024gaussianavatars}, RGBAvatar~\cite{li2025rgbavatar}, and GEM~\cite{zielonka2025gem}. 
GaussianAvatars optimizes Gaussian splats that are rigged to a parametric morphable face model against multi-view videos. 
RGBAvatar follows a similar approach, yet it also estimates Gaussian blendshapes from the face model parameters that can model dynamic appearance and geometry changes beyond the underlying face model. 
GEM first optimizes a CNN-based high-quality head avatar, which is then distilled into a lightweight, blendshape-based representation that can be directly animated from driving images.

GaussianAvatars and RGBAvatar can be directly driven with parameters of the FLAME 3DMM. 
For image-based animation, we estimate these parameters with EMOCA~\cite{danvevcek2022emoca}, a state-of-the-art 3DMM estimation method, which is also used as a pretrained feature extractor to drive GEM and our method. 
We found it beneficial for RGBAvatar to also use the EMOCA predictions during training.
The performances for self- and cross-reenactment are evaluated on five subjects from the Ava-256 dataset.
All methods are trained on a subset of the available sequences, avoiding extreme head and shoulder movements, protruded tongues, and isolated eye movements with a neutral face, while leaving out the \texttt{EXP\_free\_face} sequence for validation. 
Since we only aim to extract facial expressions from the driving image, not the global rigid transformation, we use the ground truth global pose from the VHAP tracking for the baselines and from \modelname's registrations for our method during evaluation.

\subsection{Detailed Reconstruction Time Analysis}
\label{sec:supp_avatar_timing}
\begin{table*} 
\centering
\resizebox{\linewidth}{!}{%
\begin{tabular}{l|r|r|r|r}
\toprule 
Method & GA~\cite{qian2024gaussianavatars} & RGBAvatar~\cite{li2025rgbavatar} & GEM~\cite{zielonka2025gem} & Ours\\
\midrule
Stage-Wise Durations &
\begin{tabular}[r]{@{}r@{}}VHAP Tracking: 10.65h \\ Avatar Optimization: 4.83h\end{tabular}
&
\begin{tabular}[r]{@{}r@{}}VHAP Tracking: 10.65h \\ Avatar Optimization: 0.75h\end{tabular}
&
\begin{tabular}[r]{@{}r@{}}VHAP Tracking: 10.65h \\ CNN-Avatar Optimization: 27.70h\\ Regressor Training: 6.94h\end{tabular}
&
\begin{tabular}[r]{@{}r@{}}Coarse Mesh Registration: 0.09h\\ \modelname Inference: 0.44h\\ Canonicalization \& PCA Decomp.: 0.16h \\ Emoca \& Deca Inference: 0.05h\\ PCA Refinement: 2.75h \\ Expression Regressor Training: 1.14h\end{tabular}
\\
\midrule
Total Reconstruction Time & 15.48h & 11.40h & 45.29h & 4.63 h\\
\bottomrule
\end{tabular}
}
\caption{Head avatar reconstruction time breakdown. The measurements were conducted on a representative training sequence with 3,212 frames.}
\label{tab:supp_timing}
\end{table*}
\Cref{tab:supp_timing} presents a detailed breakdown of the time cost distribution across the individual stages of head avatar reconstruction for each method.
The measurements were taken on a representative training sequence with 3212 frames using a compute node with a single NVIDIA A100 40GB GPU, 16 CPUs, and 500GB of RAM. 
To ensure full GPU usage during VHAP tracking, we ran two processes in parallel.
File system operations, e.g., data loading and writing, were excluded from all timing computations since they are highly system-dependent. 
We find that the major bottleneck of the baseline's reconstruction time, especially for RGBAvatar, lies in the multi-view head tracking. While RGBAvatar reports an impressive reconstruction time of only $80s$ for the monocular setting, in the multi-view setting, they require optimization-based tracking with VHAP~\cite{qian2024vhap}, which takes 10.65h on a representative 3212 frame training sequence, while the avatar optimization time increases to 0.75h\footnote{Experiments conducted with hyperparameters from the official code base.}. 
GEM's multi-stage approach of first tracking a parametric head model, then optimizing a high-quality head avatar, followed by a distillation, even increases the total reconstruction time per avatar to 45.3h in our setup. 
Instead, \modelname allows for skipping the lengthy optimization-based mesh registration by directly predicting registered Gaussians from the multi-view images, which takes 0.53h for the entire training sequence compared to 10.65h of optimization-based tracking with VHAP.
Unposing and PCA decomposition take 0.16 hours, such that we can start the refinement of the blendshapes and training of the expression regressor even before any of the baselines has completed registering just one 10th of the training frames.

\subsection{Further qualitative comparisons}
\begin{figure*}[t]
    \centering
    \def\svgwidth{.85\linewidth}
    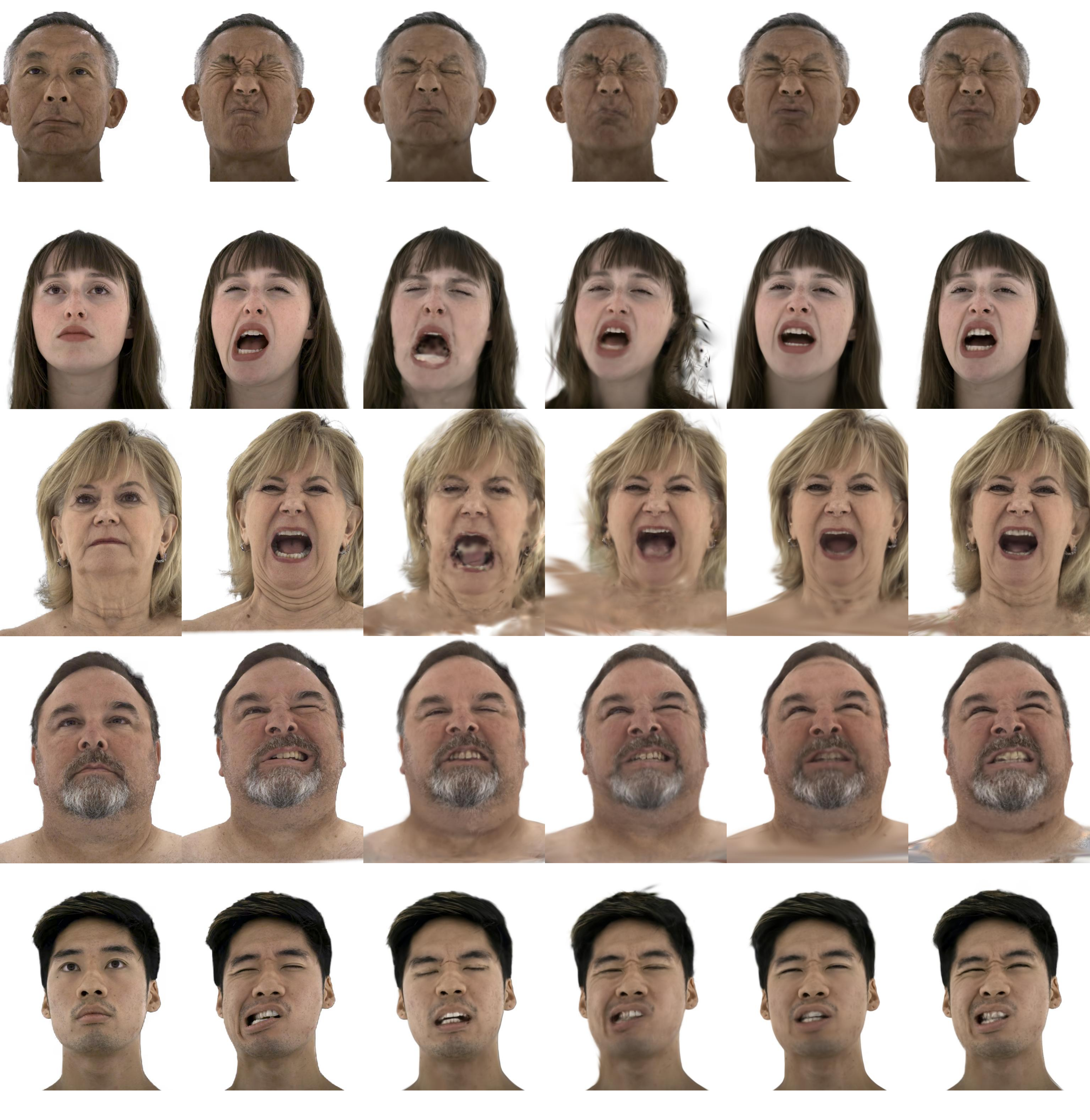
    \caption{Additional self-reenactment results for the personalized head avatars. }
    \label{fig:avatars_qual_big_test}
\end{figure*}

\begin{figure*}[t]
    \centering
    \def\svgwidth{.85\linewidth}
    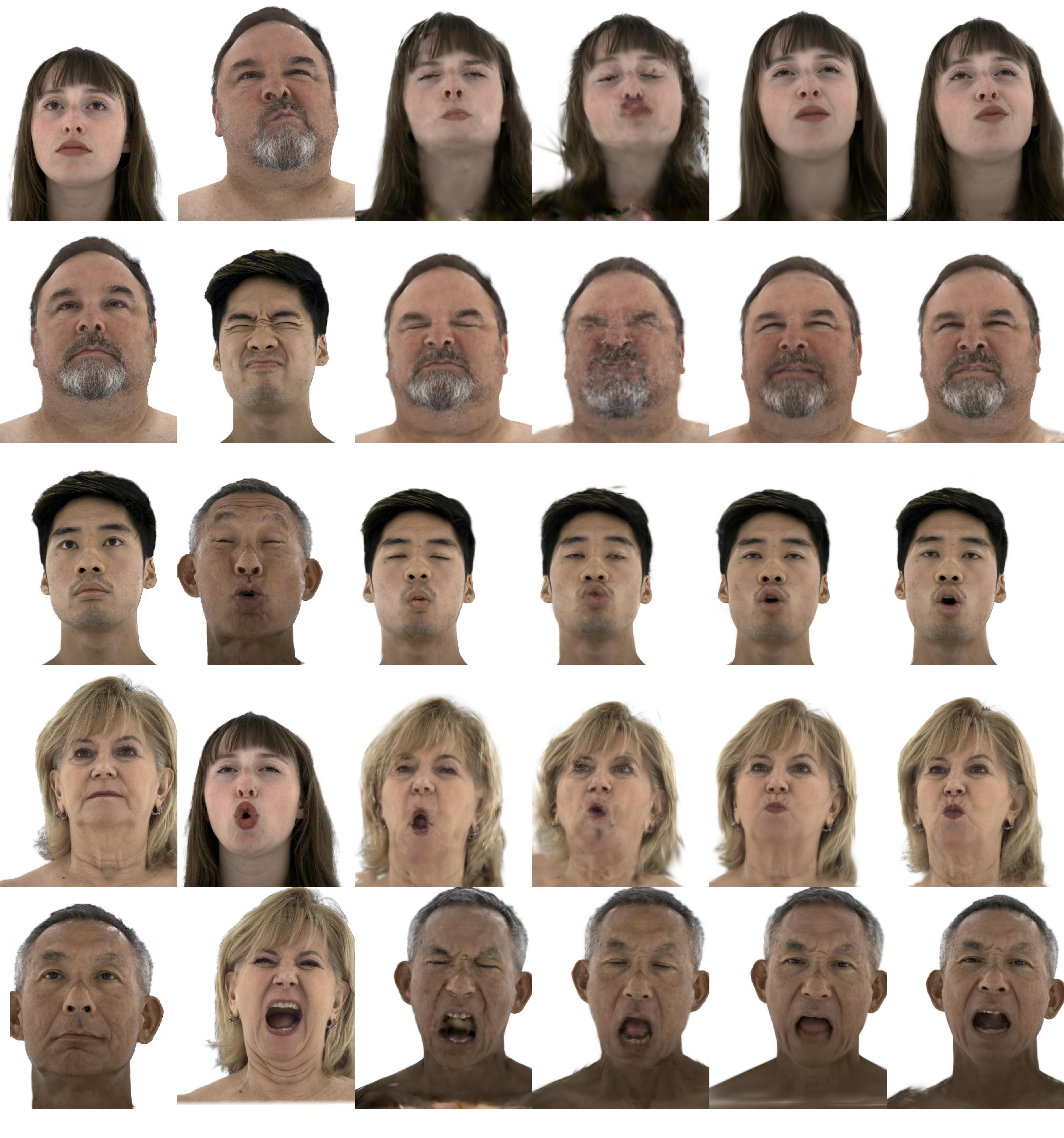
    \caption{Additional cross-reenactment results for the personalized head avatars. }
    \label{fig:avatars_qual_big_cross}
\end{figure*}

\Cref{fig:avatars_qual_big_test} and \Cref{fig:avatars_qual_big_cross} present further results of the personalized head avatars for self- and cross-reenactment respectively.

\subsection{Ablation Study}
\label{sec:supp_avatar_ablation}
\label{sec:supp_dist_ablations}
\begin{figure*}[!t]
    \centering
    \def\svgwidth{\linewidth}
    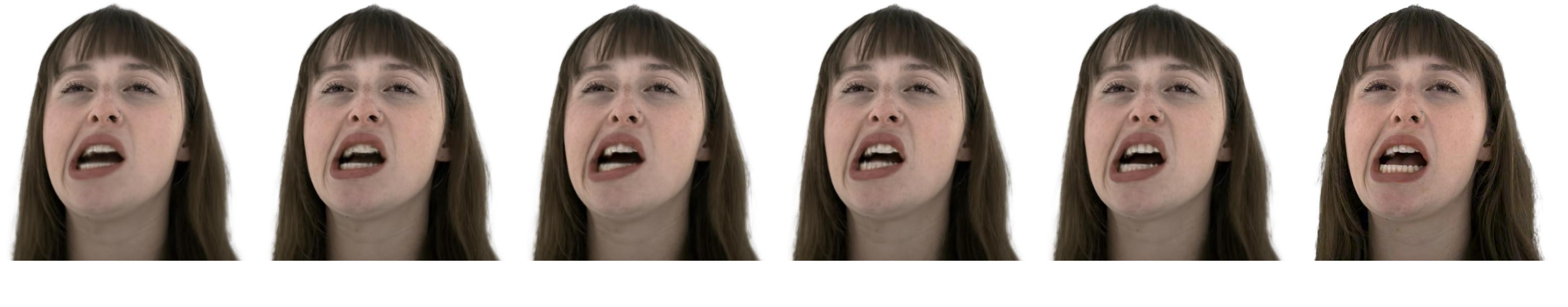
    \caption{Qualitative ablation study of the changes applied to GEM~\cite{zielonka2025gem} to create subject-specific head avatars from \modelname's predictions. Ours uses 150 PCA components.}
    \label{fig:dist_ablation}
\end{figure*}

\begin{table}   %
\centering
\resizebox{\linewidth}{!}{%
\begin{tabular}{l|rrrrr|rr}
\toprule 
& \multicolumn{5}{c|}{Self-Reenactment} & \multicolumn{2}{c}{Cross-Reenactment}\\
& LPIPS $\downarrow$&CSIM $\uparrow$&SSIM $\uparrow$&L1 $\downarrow$&PSNR $\uparrow$&CSIM $\uparrow$&EmoL1 $\downarrow$\\
\midrule
Modality-Specific PCAs & 0.180 & 0.879 & \cellcolor{tabfirst}0.816 & 0.027 & \cellcolor{tabfirst}24.376 & \cellcolor{tabfirst}0.815 & 9.849 \\
Dynamic Mouth & \cellcolor{tabfirst}0.174 & 0.878 & \cellcolor{tabsecond}0.809 & 0.027 & 24.112 & 0.811 & \cellcolor{tabfirst}9.792 \\
\# PCA Comp. = 50 & 0.177 & 0.876 & 0.808 & 0.027 & \cellcolor{tabsecond}24.126 & \cellcolor{tabsecond}0.814 & 9.891 \\
\# PCA Comp. = 100 & 0.175 & \cellcolor{tabfirst}0.880 & \cellcolor{tabsecond} 0.809 & 0.027 & 24.112 & 0.814 & 9.907 \\
Ours & \cellcolor{tabfirst}0.174 & \cellcolor{tabfirst}0.880 & \cellcolor{tabsecond} 0.809 & 0.027 & 24.122 & 0.813 & \cellcolor{tabsecond}9.837 \\
\bottomrule
\end{tabular}
}
\caption{Quantitative ablation study of the changes applied to GEM~\cite{zielonka2025gem} to create subject-specific head avatars from \modelname's predictions. By default, we use 150 PCA components.}
\label{tab:distillation_ablation}
\end{table}

\Cref{fig:dist_ablation} and \Cref{tab:distillation_ablation} present qualitative and quantitative ablation studies of the changes applied to GEM~\cite{zielonka2025gem} to create subject-specific head avatars from \modelname's predictions.
We find that employing separate PCAs for the individual Gaussian modalities ('Modality-Specific PCAs'), i.e., location, color, scale, rotation, and opacity, yields inferior results compared to jointly modeling all attributes in a single PCA.
This aligns with the intuition that the different Gaussian attributes are highly correlated (e.g., raising the eyebrows results in darker colors for wrinkles on the forehead). 
Modeling the mouth interior with Gaussians with dynamically changing color, opacity, and scale ('Dynamic Mouth') does not change the quantitative scores significantly, yet slightly reduces the faithfulness of extreme expressions at test time (see \Cref{fig:dist_ablation}). We deduce that the lightweight image-based expression encoder fails to learn the intricate dynamics of the highly dynamic mouth interior Gaussians and benefits from additional consistency constraints enforced through static colors, opacities, and scales in this region. 
Increasing the number of PCA components improves the perceptual quality by adding high-frequency details. Note that even with the highest number of PCA components that we test, i.e., our default value of 150, we still use fewer components than GEM's modality-specific PCAs with a total of 180 components. 

\section{Ethical Considerations}
\label{sec:supp_ethics}
Our method relies on multi-view studio captures with calibrated cameras, ensuring that all participants were aware of and consented to data collection. 
However, with the emergence of generative multi-view models such as CAP4D~\cite{taubner2025cap4d}, similar data could be fabricated synthetically. 
This raises potential ethical concerns regarding consent and misuse, which we strongly discourage.


\end{document}